\documentclass[9pt,twocolumn]{extarticle}
\usepackage[a4paper,top=17mm,bottom=18mm,left=15mm,right=15mm,columnsep=6mm]{geometry}
\usepackage[T1]{fontenc}\usepackage{amsthm}\usepackage{amsmath}\usepackage{newtxtext}\usepackage{newtxmath}
\usepackage[scaled=0.92]{helvet}
\usepackage{graphicx}
\usepackage{xcolor}\usepackage{textcomp}\usepackage{booktabs,array}\usepackage{placeins}\usepackage{pifont}\usepackage{enumitem}
\usepackage[numbers,super,sort&compress]{natbib}\setcitestyle{super,comma,open={},close={}}
\usepackage[hidelinks]{hyperref}\hypersetup{colorlinks=true,urlcolor=blue,linkcolor=black,citecolor=black}
\let\origFloatBarrier\FloatBarrier\renewcommand{\FloatBarrier}{}
\usepackage{caption}\DeclareCaptionLabelSeparator{pipe}{ $|$ }
\usepackage{titlesec}
\titleformat{\section}{\sffamily\bfseries\fontsize{11}{13}\selectfont}{}{0pt}{}
\titlespacing*{\section}{0pt}{9pt}{3pt}
\titleformat{\subsection}{\sffamily\bfseries\fontsize{9}{11}\selectfont}{}{0pt}{}
\titlespacing*{\subsection}{0pt}{7pt}{2pt}
\titleformat{\subsubsection}{\sffamily\bfseries\itshape\fontsize{8.5}{10}\selectfont}{}{0pt}{}
\titlespacing*{\subsubsection}{0pt}{5pt}{1pt}
\newtheoremstyle{nat}{4pt}{4pt}{\itshape}{}{\sffamily\bfseries}{.}{0.5em}{}
\newtheoremstyle{natdef}{4pt}{4pt}{}{}{\sffamily\bfseries}{.}{0.5em}{}
\theoremstyle{nat}\newtheorem{proposition}{Proposition}
\theoremstyle{natdef}\newtheorem{definition}{Definition}
\newcommand{\runhead}[1]{\textit{#1}\ }
\newcommand{\bmhead}[1]{\par\smallskip\noindent\textsf{\textbf{#1}}\ }

\newcommand{\R}{\mathbb{R}}\newcommand{\x}{\mathbf{x}}\newcommand{\y}{\mathbf{y}}\newcommand{\DLP}{\mathrm{DLP}}

\makeatletter
\renewenvironment{figure}[1][tp]{\@dblfloat{figure}[tp]}{\end@dblfloat}
\@namedef{au@bohm2022}{B\"ohm et al.}
\@namedef{au@huang2022}{Huang et al.}
\@namedef{au@kobak2019}{Kobak and Berens}
\@namedef{au@kobak2021}{Kobak and Linderman}
\@namedef{au@lee2009}{Lee and Verleysen}
\@namedef{au@lee2015}{Lee et al.}
\@namedef{au@lueks2011}{Lueks et al.}
\@namedef{au@wang2021}{Wang et al.}
\renewcommand{\citet}[1]{\def\@ct@sep{}\@for\@ct@k:=#1\do{\@ct@sep\@nameuse{au@\@ct@k}\def\@ct@sep{ and }}\cite{#1}}

\makeatother
\makeatletter\newcommand{\sitablesbegin}{\onecolumn\renewenvironment{table}[1][!htbp]{\@float{table}[##1]}{\end@float}}\makeatother\newcommand{\sitablesend}{}
\begin{document}
\twocolumn[{%
\noindent{\sffamily\footnotesize\bfseries Article}\par\vspace{5mm}
\noindent{\sffamily\bfseries\fontsize{17.5}{21}\selectfont \mbox{Quantifying local and global fidelity of nonlinear embeddings}\par}\vspace{4mm}
\noindent{\sffamily\fontsize{10}{12}\selectfont Henry Han}\par\vspace{1mm}
\noindent{\sffamily\scriptsize Data Science and AI Innovation Laboratory, Baylor University, Waco, Texas, USA.\quad e-mail: Henry\_Han@baylor.edu}\par\vspace{5mm}
\noindent\begin{minipage}{\textwidth}\bfseries\fontsize{9.5}{12.2}\selectfont Dimension reduction rests on a claim: t-SNE and UMAP preserve local structure, PCA global structure. Our DLP curve is the share of each point's $k$ nearest neighbors a map keeps. Corrected for chance, it gives $L$ at $k=10$ and $G$ over 30 to 50\% of the data, local and global fidelity. At the midpoint the curve is provably a rank correlation of distances. The LG reversal test decides, with a stated error rate, which of two maps is local and which global. On twelve datasets of up to 1.38 million points, $G$ agrees with established global scores at 0.84 and 0.93, and the area under $R_{NX}$ at 0.22 and 0.30. The entity fidelity $E$, read at each point's entity size, predicts which entities a map loses beyond $L$ and $G$ where entity sizes differ. In an agent memory of one million passages, t-SNE and PCA cross at 150,000 passages ($p=2.6\times10^{-5}$).\end{minipage}\par\vspace{6mm}
}]

Embeddings are how people and machines look at large high-dimensional data. Analysts draw two-dimensional maps of single-cell atlases, pathology slides, customer records and text collections with t-SNE \citep{vandermaaten2008}, UMAP \citep{mcinnes2018}, PHATE \citep{moon2019}, TriMAP \citep{amid2019}, PaCMAP \citep{wang2021} and PCA. AI agents keep their memories as embedding vectors and retrieve them by nearest-neighbor search \citep{lewis2020rag}. Large memories are compressed, by truncation \citep{kusupati2022matryoshka}, by principal components, or by product quantization \citep{jegou2011pq}, so that they fit in memory and can be sent between agents. Both kinds of embedding are described in two words. A method preserves \emph{local} structure if nearby points stay nearby, and \emph{global} structure if the layout of clusters, branches and regions is kept. PCA is called global and t-SNE local, and UMAP is said to be more global than t-SNE. A compressed memory is judged by the recall of its nearest neighbors \citep{aumuller2020annbench}, a local score. Agents also consolidate, delete and summarize memories by topic \citep{edge2024graphrag}, and these tasks depend on global structure.

\runhead{Two questions decide which map to use.} How much local and how much global structure does a map keep? And on given data, is one map more local and another more global? The answers set the method, the parameter or the compression that an analyst or an agent designer picks. A pathologist who searches the right map for tiles like a tumor tile finds their true neighbors. An agent that groups topics on the right map keeps distant topics apart. Answering the two questions takes two things. The first is a common scale. Local structure is scored by neighbor overlap or trustworthiness \citep{venna2001}, and global structure by distance rank correlation, random-triplet accuracy \citep{wang2021} or centroid distances \citep{kobak2019}. These scores have different ranges and chance levels. The second is a test that separates a real difference from chance. PCA's global lead over t-SNE is 0.056 on one million text passages, at perplexity 200, and on Renal-Cortex, a kidney single-cell atlas, at perplexity 30. A test shows it is real on the first ($p=8.6\times10^{-6}$) and within chance on the second ($p=0.15$). The questions are now pressing, because single-cell atlases and agent memories hold millions of points, and maps of that size are judged by numbers.

\runhead{One curve gives both fidelities.} We read both kinds of fidelity from the two ends of one chance-corrected curve, the locality profile, and we compare two maps point by point. The share of each point's $k$ nearest neighbors that a map keeps, corrected for chance, is a curve over $k$. At $k=10$ it gives the local fidelity $L$. At neighborhoods of 30 to 50 percent of the data it gives the global fidelity $G$. We prove that this upper end is the agreement on which half of the data is near each point, which converges to a rank correlation of the distances. $L$ and $G$ are one statistic at two scales, so they can be compared directly. Both are means over points, so two maps can be tested against each other. This curve is our DLP curve \citep{han2022}, corrected for chance.

\begin{figure}[!htbp]
\centering
\includegraphics[width=\textwidth]{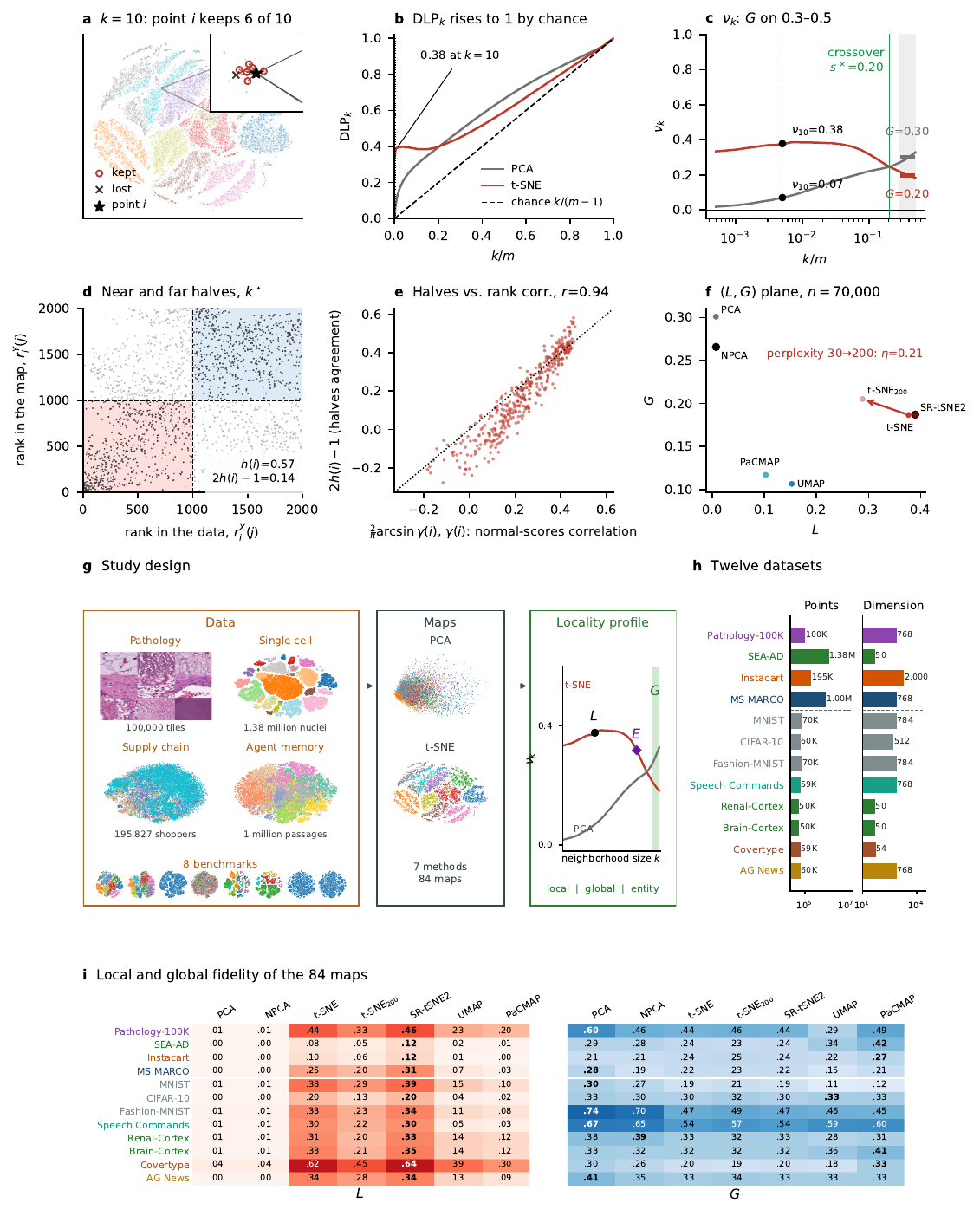}
\caption{From the locality profile to $L$ and $G$ on MNIST ($n=70{,}000$): t-SNE is more faithful below 20 percent of the data and PCA above it. (a) Point $i$ keeps 6 of its 10 nearest neighbors in the t-SNE map. (b)--(e) A subsample of $m=2{,}000$ points. (b) $\DLP_k$ rises to 1, partly by chance (dashed). (c) After the chance correction the profiles cross at $s^\times=0.20$; the shaded band gives $G$. Values in (c) come from the subsample; $L$ in the text is computed on all points. (d) Ranks from point $i$ in the data and the map; $h(i)$ is the share in the shaded quadrants. (e) $2h(i)-1$ tracks $\frac{2}{\pi}\arcsin\gamma(i)$, where $\gamma(i)$ is the normal-scores correlation of the distances from point $i$, a rank-based estimate of the Gaussian-copula correlation (Proposition~\ref{prop:sheppard}). (f) Seven maps in the $(L,G)$ plane; $\eta$ from one run, 0.33 as a three-run mean on 60,000 digits (Supplementary Fig.~\ref{fig:tradeoff}) (NPCA: nonnegative PCA; SR-tSNE2: our rank refinement of t-SNE). (g) Study design: four applications and eight benchmarks, seven maps each, scored by $L$, $G$ and $E$. (h) The twelve large datasets, 50,000 to 1.38 million points in 50 to 2,000 dimensions. (i) $L$ and $G$ of the 84 maps; bold, the highest on each dataset.}\label{fig:concept}
\end{figure}

\runhead{On MNIST \citep{lecun1998gradient} t-SNE is more faithful below 20 percent of the data and PCA above it.} The profile starts from the degree of locality preservation, $\DLP_k$, the average fraction of each point's $k$ nearest neighbors that remain among its $k$ nearest in the map \citep{han2022}. Read over all $k$ and corrected for chance, it gives the locality profile. In the t-SNE map, point $i$ keeps 6 of its 10 nearest neighbors (Figure~\ref{fig:concept}a). Averaged and corrected, this overlap gives the profile (Figure~\ref{fig:concept}b,c). At the upper end, each point's agreement with the data on its near and far halves tracks the rank correlation of its distances. The Pearson correlation is 0.94 over 400 points (Figure~\ref{fig:concept}d,e). PCA overtakes t-SNE at $s^\times=0.20$, the crossover scale. Raising t-SNE's perplexity from 30 to 200 gains 0.21 of global fidelity for each unit of local fidelity traded, the exchange rate $\eta$ (Figure~\ref{fig:concept}c,f).

\runhead{Our DLP curve \citep{han2022} decides between two maps.} Quality curves have so far scored maps one at a time, with no test between two. Three results make the test possible. First, a proof: at its midpoint the curve tends to a rank correlation of the distances, so its upper end, read as $G$, measures global structure. Second, a curve for every point: each score becomes a mean over points with a standard error, which gives the LG reversal test and an interval for the size at which to change maps. Third, a reading at each point's entity size: the entity fidelity $E$ tells which entities a map loses. None of the closest methods \citep{lee2009,lueks2011,lee2015,kury2026,wang2021,kobak2019} gives the upper end of a neighbor curve a proven meaning or tests whether two maps reverse their order. The area under $R_{NX}$ agrees with established global scores at only 0.22 and 0.30 over the 84 default maps, against 0.84 and 0.93 for $G$ (Supplementary Table~\ref{tab:novelty}).

\begin{figure}[p]
\centering
\includegraphics[width=0.88\textwidth]{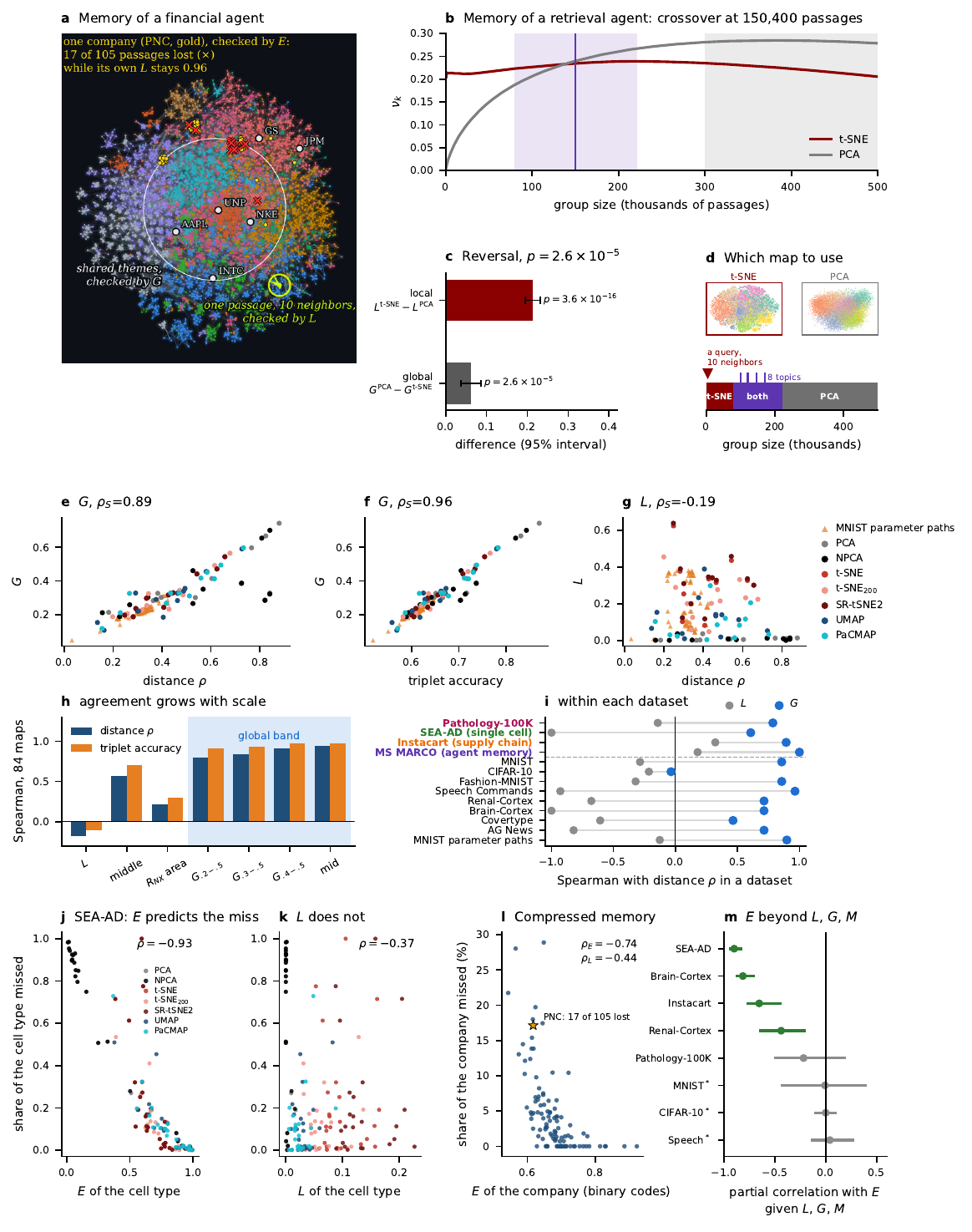}
\caption{$L$ and $G$ tell which map to use for a memory, and $E$ tells which entities a map or a compressed memory loses. (a) The memory of a financial agent on an SR-tSNE2 map: 8,356 passages of 106 S\&P 500 companies, colored by theme. After binary compression and rescoring, PNC Financial Services (PNC, gold) keeps $L=0.96$ and loses 17 of its 105 passages ($\times$); circles mark one passage with its 10 neighbors ($L$) and the shared themes ($G$). (b)--(d) MS MARCO: t-SNE for the passages near a query, PCA for the layout of topics, crossover at 150,400 passages. (b) Profiles; purple line and band, the crossover and its 95\% interval; gray band, the range that gives $G$. (c) The two one-sided tests with 95\% intervals. (d) Which map answers each question. (e)--(i) $G$ measures what the established global scores measure. (e)--(g) $G$ and $L$ against the distance rank correlation and random-triplet accuracy over 156 maps. (h) Agreement rises with the scale at which the profile is read. (i) Within each dataset, $G$ orders the methods as the distance rank correlation does. (j)--(m) Entity fidelity predicts which entities a map or a compressed memory loses, beyond $L$, $G$ and a common scale. (j) SEA-AD: miss rate of each cell type against its $E$, 24 cell types on seven maps. (k) The same pairs against $L$. (l) The same memory compressed to binary codes: miss rate of each company after rescoring, against $E$ of the company on the codes; star, PNC. (m) Partial rank correlation of the miss rate with $E$ given $L$, $G$ and $M$, the profile read at the median entity size, with 95\% bootstrap intervals over entities; green, interval below 0; $^*$, entities of nearly equal size.}\label{fig:agentmem}
\end{figure}

\runhead{The LG reversal test tells which map to use.} It decides with a stated error rate whether one map is local and another global. It is designed for any local and any global dimension reduction method, because it needs the data and the two maps alone. It runs two one-sided tests on the scores of single points, one at each end of the profile. A block jackknife accounts for the dependence between neighboring points, and the intersection--union principle combines the two tests into one decision. Figure~\ref{fig:agentmem}a shows what the two scores check in the memory of a financial agent. A recall check sees one passage and its 10 neighbors, and a topic check sees the shared themes. Figure~\ref{fig:agentmem}b--d applies the test to the memory of a retrieval agent: one million MS MARCO passages \citep{bajaj2016msmarco} embedded with BGE-base \citep{xiao2024cpack}. t-SNE is more local than PCA by 0.213 and PCA is more global by 0.062, and the test declares the reversal with $p=2.6\times10^{-5}$. The crossover lies at 150,400 passages, with 95\% interval 79,100 to 221,700. Retrieval itself runs in the full embedding space; the maps serve the person who audits the memory. That person should use the t-SNE map to inspect the passages near a query and any subtopic of fewer than 79,000 passages, and the PCA map to see which of the eight $k$-means topics lie far apart and which to merge or delete.

\runhead{Contributions.} We make three contributions. First, from our DLP curve \citep{han2022} we derive the local fidelity $L$, the global fidelity $G$, whose upper end has a proven meaning, and the entity fidelity $E$. Against an uncompressed reference, $\DLP_k$ is the recall at $k$ by which vector search is already judged, so $G$ and $E$ measure what recall cannot see. Second, the LG reversal test decides, with a stated error rate, whether one map is local and another global, and returns, with a 95\% interval, the size at which to change maps. Third, the scores change what applications conclude. The claim that t-SNE is local and PCA global holds firmly on six of twelve large datasets, narrowly on three, within chance on two, and is reversed on one. In a compressed financial memory, PNC Financial Services keeps $L=0.96$ yet loses 17 of its 105 passages, and $E$, read before any query, flags such companies. In a memory of one million passages, the test places the change from t-SNE to PCA near 150,000 passages. The released SEA-AD UMAP \citep{gabitto2024seaad} is more faithful than PCA only below 16\% of the data.

\section{Results}\label{sec:results}

\subsection{The LG reversal test: which map answers which question}\label{sec:lgtest}

\runhead{Each comparison returns a declaration, a p-value and a crossover.} For two maps $A$ and $B$ of the same data, the test returns one of three declarations. Either $A$ is local and $B$ global, or $B$ is local and $A$ global, or one map is at least as faithful at both ends. Then the user needs that one map. With a reversal it returns a p-value and the crossover size $s^\times n$ with its 95\% interval. A question about fewer than $s^\times n$ points goes to the local map, and a question about more points goes to the global map. Figure~\ref{fig:agentmem}b--d works through one example.

\subsection{The upper end of the profile measures global structure}\label{sec:validation}

The test relies on $G$ measuring global structure. $G$ ranks maps as the established global scores do, and it is stable across bands and subsamples (Supplementary Note~\ref{app:sensitivity}).

\runhead{$G$ agrees with the established global scores.} Over the 84 maps of the twelve large datasets, $G$ has rank correlation 0.84 with the distance rank correlation and 0.93 with random-triplet accuracy. Adding the 72 maps of the MNIST parameter paths, it reaches 0.89 and 0.96 (Figure~\ref{fig:agentmem}e,f). These are the two scores that \citet{kobak2019} and \citet{wang2021} used for global structure, and they correlate with each other at 0.97. $L$ correlates with them at $-0.19$ (Figure~\ref{fig:agentmem}g), so it carries separate information. On large data the maps that keep neighbors best and the maps that keep the global layout best are different maps. Correlation rises with the scale at which the profile is read (Supplementary Table~\ref{tab:validation} and Figure~\ref{fig:agentmem}h). The area under $R_{NX}$, the usual summary of neighbor curves \citep{lee2015}, reaches 0.22 and 0.30, because it weights the lower end.

\runhead{Within each dataset $G$ orders the methods as the distance rank correlation does.} The rank correlation is 0.71 or more on ten of the thirteen groups of maps, the twelve datasets and the MNIST paths, with a median of 0.79 (Figure~\ref{fig:agentmem}i). It exceeds that of $L$ on all thirteen. Proposition~\ref{prop:sheppard} explains the agreement: the upper end of the profile is a rank correlation of distances, read point by point.

\subsection{Entity fidelity predicts which entities a map loses}\label{sec:entity}

Between the two ends lies the scale of an entity, and there $E$ predicts which entities a map loses. The entity fidelity $E$ reads the profile at the size of each point's own entity, such as its cell type (Methods). $M$ reads it at the median entity size for every point.

\runhead{Where entity sizes differ, $E$ predicts the miss rate beyond $L$, $G$ and $M$.} The partial correlation is $-0.90$ on SEA-AD, whose 24 cell types hold 1,496 to 330,085 nuclei (95\% interval $-0.94$ to $-0.83$; Figure~\ref{fig:agentmem}j,m). On the same cell types and maps the rank correlation of the miss rate is $-0.93$ with $E$ and $-0.37$ with $L$ (Figure~\ref{fig:agentmem}k). The partial correlation is $-0.81$ on Brain-Cortex ($-0.87$ to $-0.71$), $-0.65$ on Instacart ($-0.77$ to $-0.45$) and $-0.44$ on Renal-Cortex ($-0.64$ to $-0.21$). Where entities have nearly equal sizes, $E$ and $M$ read nearly the same $k$ and carry the same information (Figure~\ref{fig:agentmem}m).

\runhead{$E$ flags the companies that a compressed memory loses.} An agent's memory of 8,356 passages from the annual reports of 106 S\&P 500 companies \citep{chen2021finqa}, embedded with BGE-base \citep{xiao2024cpack}, was compressed to binary codes. Each company's passages were selected as a vector database does: twice as many candidates as passages, at least 100, were taken on the codes and rescored at full precision \citep{osquant2025}. The selection missed passages of their own company for 65 of the 106 companies, 223 passages in all. $E$, read on the codes before any query, has rank correlation $-0.74$ with the miss rate, and $L$ has $-0.44$. The partial correlation of $E$ given $L$ is $-0.66$ (95\% interval $-0.76$ to $-0.52$; Figure~\ref{fig:agentmem}l). PNC Financial Services (PNC) keeps $L=0.96$ after rescoring and loses 17 of its 105 passages (Figure~\ref{fig:agentmem}a); its $E$ is the 16th lowest of 106.

\subsection{Four applications: pathology, single cells, supply chains, and AI agent memory}\label{sec:apps}

\begin{figure}[!htbp]
\centering
\includegraphics[width=\textwidth]{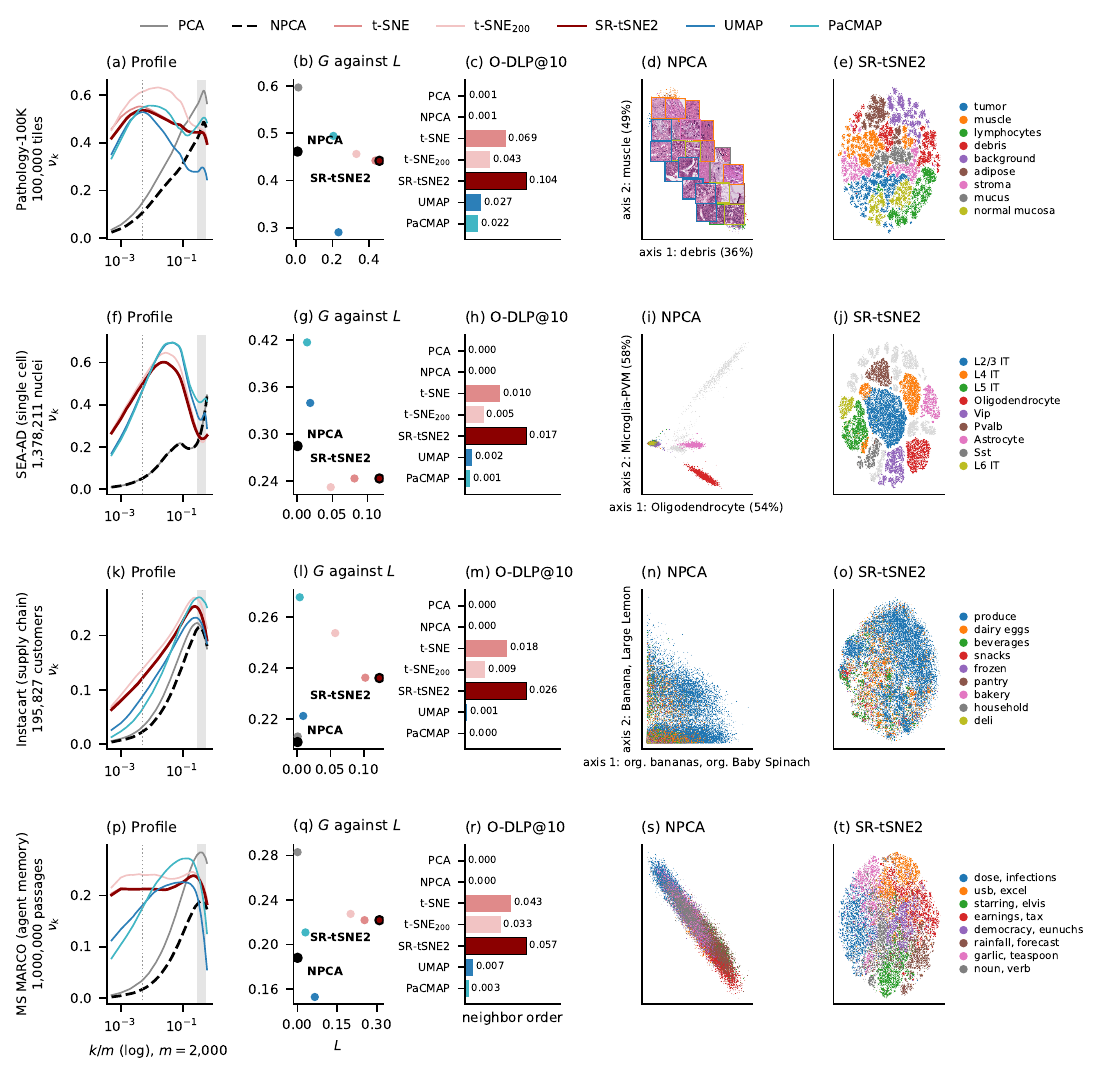}
\caption{On the four applications SR-tSNE2 is the most local map, and the most global map is PCA on two and PaCMAP on two. Rows: Pathology-100K, SEA-AD, Instacart and MS MARCO. Columns: the profile; $G$ against $L$; O-DLP@10, the share of the 10 nearest neighbors returned at their exact rank, which is what a ranked search result shows; the NPCA map with named axes, on Pathology-100K with the hematoxylin and eosin (H\&E) tile nearest each cell of a $6\times6$ grid, framed in its tissue color; the SR-tSNE2 map colored by tissue, cell subclass, department or $k$-means cluster. Maps show 20,000 random points.}\label{fig:appsbig}
\end{figure}

In all four applications $L$ and $G$ pick different maps: SR-tSNE2, our rank refinement of t-SNE, is the most local, and the most global map changes with the data. Figure~\ref{fig:appsbig} shows the four applications, and Supplementary Fig.~\ref{fig:apps} three smaller datasets. On Pathology-100K and MS MARCO PCA has the highest $G$, 0.597 and 0.283 (Figure~\ref{fig:appsbig}b,q). On SEA-AD and Instacart PaCMAP has the highest $G$, 0.417 and 0.268, and t-SNE is above PCA on Instacart, 0.236 against 0.213 (Figure~\ref{fig:appsbig}g,l). The claim that PCA is global thus holds on Pathology-100K and MS MARCO, and on SEA-AD and Instacart PaCMAP takes its place; the reversal test below checks each case, and Supplementary Table~\ref{tab:pairs} tests PaCMAP.

\runhead{The score that each task needs picks the map.} In each application some tasks depend on $L$: finding the tiles that resemble a tumor tile, the cells of a rare type, the customers who buy like a given one, and the passages near a query in an agent's memory. Other tasks depend on $G$: the layout of tissue classes, the distances between cell types, customer segmentation and stocking, and topic consolidation in an agent's memory. For the first kind, SR-tSNE2 gives the highest $L$ on all four datasets: 0.458 on Pathology-100K, 0.117 on SEA-AD, 0.124 on Instacart and 0.309 on MS MARCO.

\FloatBarrier

\FloatBarrier
\subsection{The reversal test at scale}\label{sec:lgscale}

The reversal test decides which map serves which task: the usual claim holds firmly on six of the twelve large datasets, and the crossover tells a user which map to use for each question.

\subsubsection{A worked example: the memory of a retrieval agent}\label{sec:memexample}

For the memory of a retrieval agent the test declares t-SNE local and PCA global, and it places the change of maps near 150,000 passages.

\runhead{Both one-sided tests reject.} We test t-SNE against PCA on a uniform subsample of 5,000 passages with 20 $k$-means blocks (Figure~\ref{fig:agentmem}b--d). The local test gives $L^{\rm t\text{-}SNE}-L^{\rm PCA}=0.213$ with jackknife standard error 0.009 and $p_L=3.6\times10^{-16}$. The global test gives $G^{\rm PCA}-G^{\rm t\text{-}SNE}=0.062$ with standard error 0.012 and $p_G=2.6\times10^{-5}$. Both nulls are rejected, and the reversal is declared with $p=2.6\times10^{-5}$.

\runhead{Where to change maps.} The crossover is $s^\times=0.150$, with 95\% interval 0.079 to 0.222, that is, 150,400 passages (79,100 to 221,700). Figure~\ref{fig:agentmem}d lists which map answers each question about the memory. SR-tSNE2 and UMAP give the same declaration, with crossovers at 150,200 and 135,600 passages.

\subsubsection{Seven of twelve large datasets show the reversal}

\begin{figure}[!htbp]
\centering
\includegraphics[width=\textwidth]{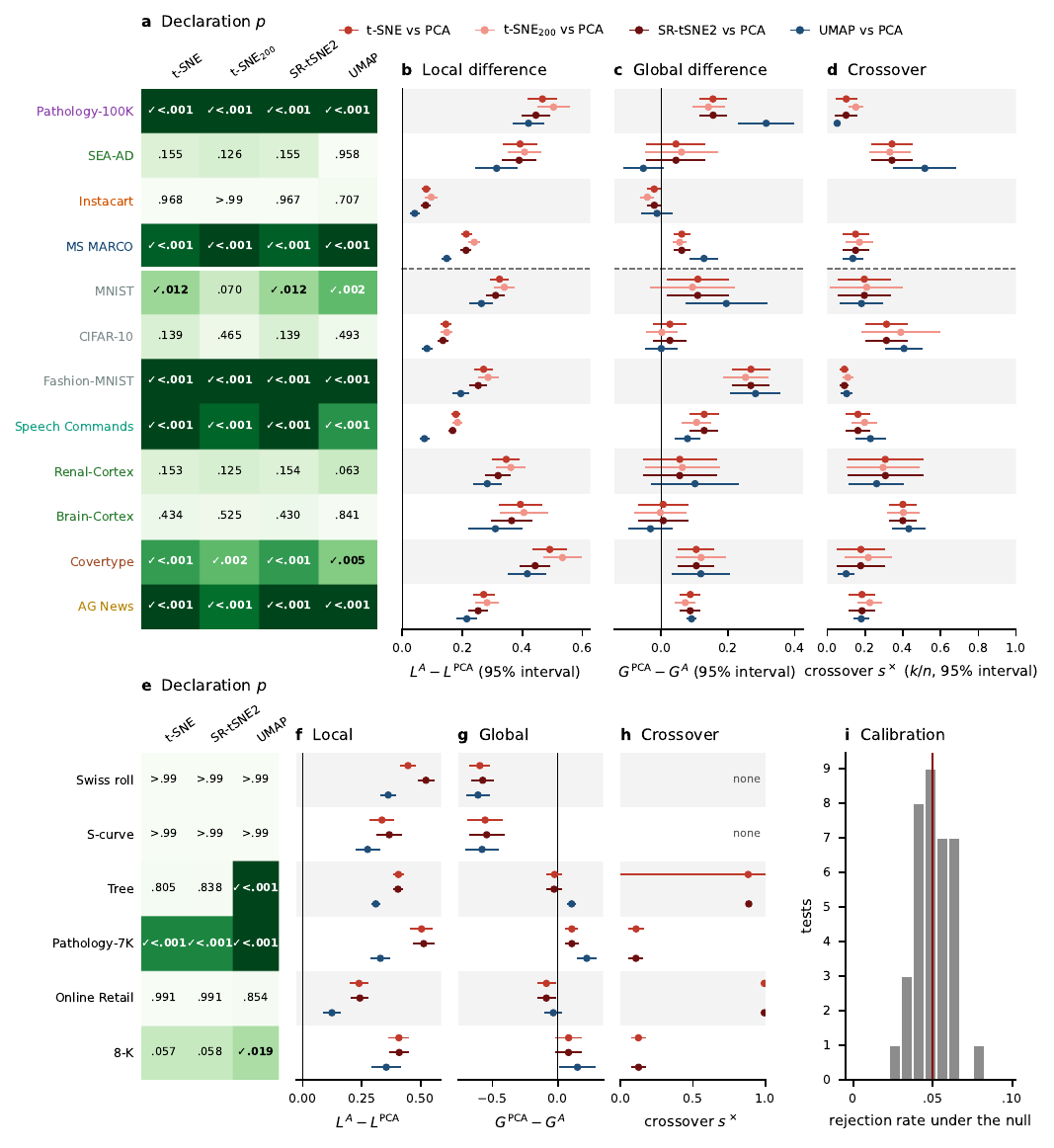}
\caption{The reversal test declares t-SNE local and PCA global on seven of the twelve large datasets at the default 20 blocks, and identifies where the neighbor embeddings (t-SNE, SR-tSNE2 and UMAP) are the more global maps. (a)--(d) Large datasets, applications above the line, four maps against PCA: (a) p-value of the declaration, check mark at level 0.05; (b) $L^A-L^{\rm PCA}$ and (c) $G^{\rm PCA}-G^A$ with 95\% intervals; (d) crossover scale with its 95\% interval. (e)--(h) The same for the smaller datasets and the synthetic manifolds. (i) Rejection rates of the global test under a null that resamples the 20 blocks; all lie near the level 0.05.}\label{fig:large}
\end{figure}

Run on all twelve large datasets, the test declares t-SNE local and PCA global on seven: MNIST, Fashion-MNIST, Speech Commands, Covertype, AG News, Pathology-100K and MS MARCO, each with $p\le0.012$ (Figure~\ref{fig:large}a). UMAP and SR-tSNE2 are declared local on the same seven. Six of the seven hold with 10, 20 and 50 blocks; on these the claim holds firmly. It holds narrowly on MNIST, at $p=0.061$ with 10 blocks, and on CIFAR-10 and Renal-Cortex, declared with 50 blocks (Supplementary Note~\ref{app:blocks}). On the other five, t-SNE is more local with $p_L<10^{-4}$. The global gap is within chance on Brain-Cortex (0.006) and SEA-AD (0.044), and within its 20-block sampling error on CIFAR-10 (0.026) and Renal-Cortex (0.056). On Instacart t-SNE is the more global map, by 0.021 ($p=0.032$). For t-SNE the crossover lies at $k/n$ of 0.10 on Pathology-100K (0.04 to 0.16) and 0.15 on MS MARCO (0.08 to 0.22). On the other five datasets with a declared reversal it lies between 0.09 and 0.20 (Figure~\ref{fig:large}d).

\subsubsection{Where the neighbor embeddings are global}
On four datasets the neighbor embeddings (t-SNE, SR-tSNE2 and UMAP) are the more global maps: the Swiss roll, the S-curve and both retail datasets. The test identifies each case (Figure~\ref{fig:large}e--i). On the Swiss roll and the S-curve, where the reference is geodesic, every neighbor embedding is more global than PCA, by 0.54 to 0.61 in $G$. One neighbor embedding thus serves both ends. On the tree PCA reaches t-SNE at $k/n=0.88$. On Online Retail, t-SNE and SR-tSNE2 are more global than PCA by 0.088, and PCA reaches them at $k/n=0.99$. There t-SNE leads at both ends, as on Instacart, so the test names it the one map to use ($p=0.99$ and 0.97). On Pathology-7K the reversal is declared with $s^\times=0.11$ (0.05 to 0.16), close to its value of 0.10 on Pathology-100K. A reversal thus depends on the pair of methods, the dataset and, through the PCA map, its size, as the next subsection shows.

\subsubsection{Large data make the reversal reproducible}

On MNIST, three nested series of subsets from 2,500 to 70,000 points show how. The reversal is declared in all three series at 10,000, 40,000 and 70,000 points and in two of three at 5,000 and 20,000 (Supplementary Fig.~\ref{fig:scale}a). PCA's two axes are stable from 10,000 points (Supplementary Note~\ref{sec:pcastab}). In smaller subsets they change from draw to draw: $G$ of PCA is under 0.20 in four of the six subsets of 2,500 and 5,000 points (Supplementary Fig.~\ref{fig:scale}c). From 10,000 points on, $G$ of PCA lies between 0.296 and 0.308 in every subset, and PCA's global advantage over t-SNE lies between 0.09 and 0.13. At perplexity 200, from 20,000 points on, the advantage lies within 0.03 of its value at perplexity 30 (Supplementary Fig.~\ref{fig:scale}b). Large data make the reversal reproducible because they make the PCA map stable.

\subsection{Published cell atlases: where a released map can be trusted}\label{sec:atlas}
The same test checks the maps that consortia release with their atlases.

\begin{figure}[p]
\centering
\includegraphics[width=0.83\textwidth]{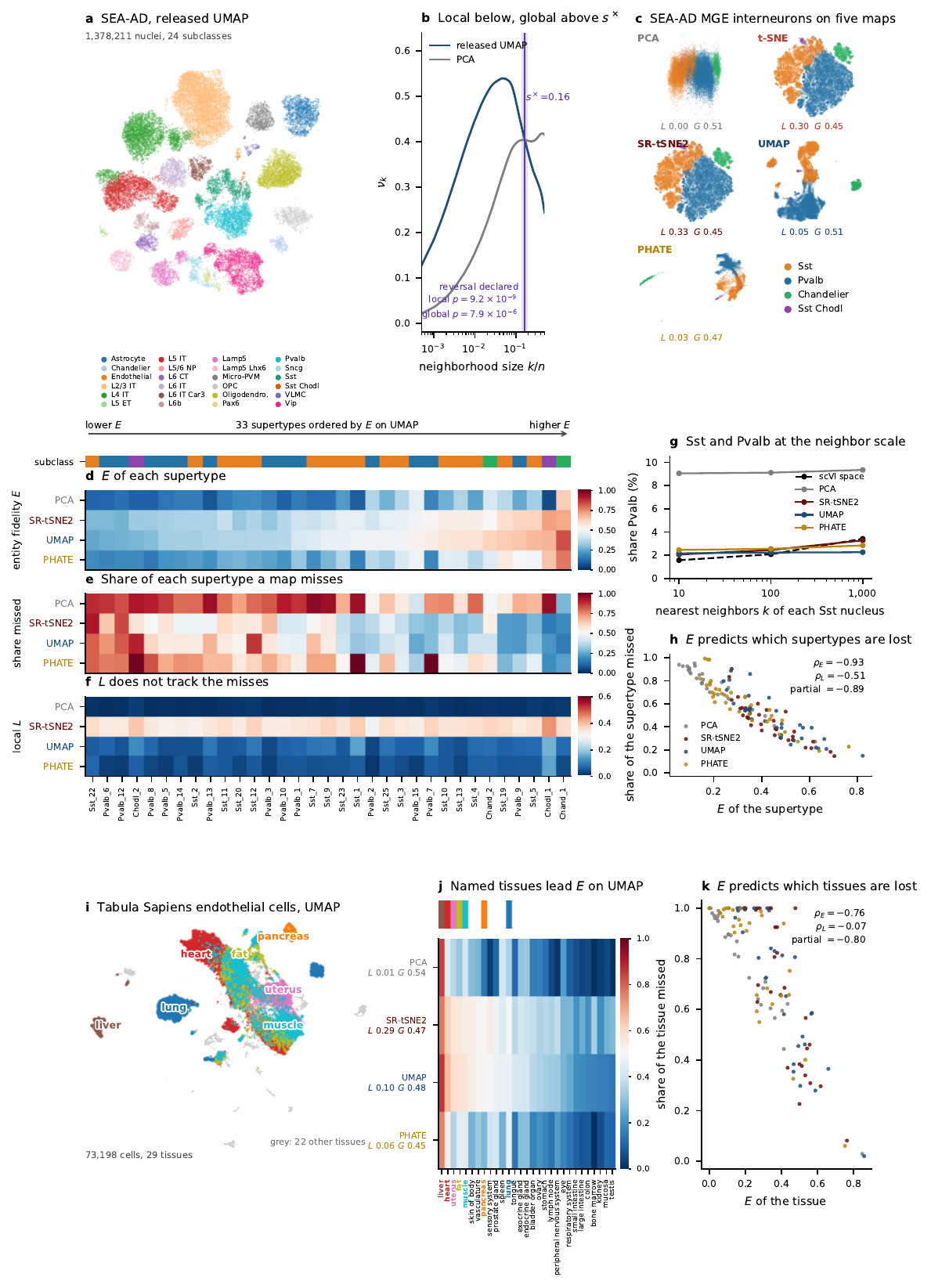}
\caption{On published cell atlases, $L$ and $G$ tell where a released map can be trusted, and $E$ tells which supertypes and tissues it loses. (a),(b) SEA-AD, the consortium's released UMAP judged against the scVI space it was built from, the latent space of a deep generative model of the counts (1,378,211 nuclei). (a) The released map, colored by subclass. (b) Locality profiles of the released UMAP and of PCA, and the two one-sided tests declare the reversal. (c)--(h) SEA-AD interneurons from the medial ganglionic eminence (MGE; 161,493 nuclei, 33 supertypes). (c) PCA, t-SNE, SR-tSNE2, UMAP and PHATE maps colored by subclass, with $L$ and $G$; against PCA the test declares a reversal for t-SNE and SR-tSNE2, and none for UMAP (global $p=0.47$) or PHATE ($p=0.079$). (d)--(f) $E$, miss rate and $L$ of each supertype on each map, ordered by $E$ on UMAP. (g) Share of Pvalb nuclei among the $k$ nearest neighbors of each Sst nucleus; UMAP keeps the two subclasses nearly as separate as the scVI space at the neighbor scale. (h) Miss rate of each supertype against its $E$. (i)--(k) Tabula Sapiens endothelial cells (73,198 cells, 29 tissues). (i) UMAP colored by tissue. (j) $E$ of each tissue on each map, ordered by $E$ on UMAP, with each map's $L$ and $G$; against PCA the test declares a reversal for every map, PHATE included (global $p=0.006$). (k) Miss rate of each tissue against its $E$.}\label{fig:atlas}
\end{figure}

\runhead{The released SEA-AD UMAP can be trusted below 16\% of the data.} The SEA-AD UMAP was computed from the scVI space \citep{lopez2018scvi}, the latent space of a deep generative model of the counts. Judged against that space, it is more faithful than PCA below $s^\times=0.16$ (95\% interval 0.135--0.193) and less faithful above it (Figure~\ref{fig:atlas}a,b; $p=9.2\times10^{-9}$ and $7.9\times10^{-6}$). Among interneurons from the medial ganglionic eminence (MGE), UMAP keeps Sst and Pvalb nuclei nearly as separate as the scVI space at the neighbor scale (2.2\% versus 1.6\% Pvalb among the 10 nearest neighbors; PCA 9.0\%; Figure~\ref{fig:atlas}c,g). On both atlases SR-tSNE2 is more local and PCA more global than PHATE~\cite{moon2019}, which is designed to keep both ($L$ 0.03 and 0.06, $G$ 0.47 and 0.45; Figure~\ref{fig:atlas}c).

\runhead{$E$ predicts which supertypes and tissues a map loses.} On SEA-AD $E$ predicts which supertypes a map loses (rank correlation $-0.93$, against $-0.51$ for $L$; Figure~\ref{fig:atlas}d--f,h). In Tabula Sapiens endothelial cells \citep{tabulasapiens2022}, it predicts which tissues a map loses ($-0.76$, against $-0.07$ for $L$; Figure~\ref{fig:atlas}i--k).

\subsection{The test on independent samples}\label{sec:simres}
\begin{figure}[!htbp]
\centering
\includegraphics[width=0.84\textwidth]{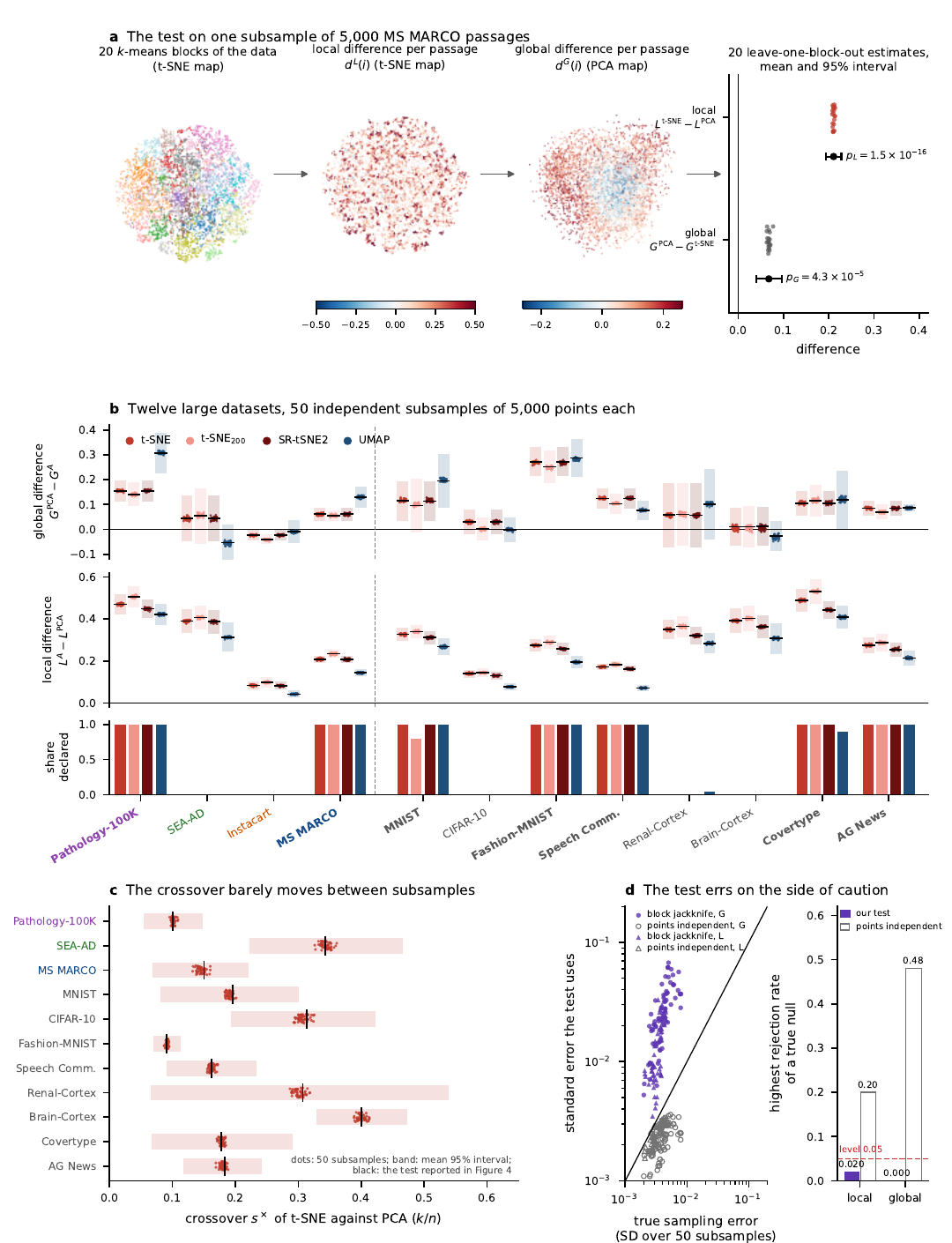}
\caption{On independent samples the LG reversal test keeps its declarations and its crossover, and it stays within its error rate. (a) The test on one uniform subsample of 5,000 MS MARCO passages, t-SNE against PCA, a different subsample from Figure~\ref{fig:agentmem}c: the 20 $k$-means blocks, the per-passage local and global differences $d^L(i)$ and $d^G(i)$, and the 20 leave-one-block-out estimates with their 95\% interval. (b) Twelve large datasets, four maps against PCA, 50 independent subsamples of 5,000 points with the maps fixed: estimates of the global and local differences (dots), the full-data global difference and the mean local difference (black), the mean jackknife 95\% interval (band), and the share of subsamples with a declared reversal; bold, datasets on which Figure~\ref{fig:large} declares a reversal for t-SNE. (c) Crossover of t-SNE against PCA in the 50 subsamples (dots), mean 95\% interval (band) and the test of Figure~\ref{fig:large} (black); on Instacart, where t-SNE leads at both ends, the profiles meet only at the end ($s^\times=0.99$). (d) Standard error used by the test against the standard deviation of the estimate over subsamples, for the block jackknife and for a test that treats points as independent, and the highest rate at which each rejects a true null at level 0.05 over six pairs of maps on the twelve datasets.}\label{fig:sim}
\end{figure}

The declarations above come from one subsample of each dataset. On 50 independent subsamples, the declarations of t-SNE against PCA and the crossovers do not change (Figure~\ref{fig:sim}). The local test keeps its error rate at or below 0.020, against up to 0.20 for a test that treats points as independent.

\FloatBarrier

\section{Discussion}\label{sec:discussion}

Local and global fidelity are one statistic read at two scales, so whether a map is local or global becomes a question with a tested answer.

\runhead{Analysts choose a map by the size of the question.} The upper end of the profile has a proven meaning: the agreement on which half of the data is near each point. Both ends are the same statistic, so the reversal test turns their comparison into a decision, and the crossover marks where to change maps. A parameter value, such as a perplexity, becomes a point $(L,G)$ on the path that the parameter traces. Across 24 parameter settings on 60,000 digits, PCA keeps the highest $G$ and t-SNE the highest $L$ (Supplementary Note~\ref{sec:findings}). A claim of global structure should report $G$ with its run-to-run spread.

\runhead{The same scale applies to the memory of an AI agent.} For a compressed agent memory, the uncompressed embedding is the natural reference. Against it, $\DLP_k$ is the recall of the compressed index at $k$, the quantity by which approximate nearest-neighbor search is already evaluated \citep{aumuller2020annbench}, and $L$ adds the correction for chance. $G$ is new to this setting. It measures whether topics, episodes or documents that lie far apart in the full memory stay far apart in the compressed one. Questions about a whole corpus depend on this organization \citep{edge2024graphrag}. $G$ can be estimated from a few thousand stored items, however large the memory (Methods). $E$ adds a check that needs no query: read on the codes of a compressed memory, it flags the companies whose passages retrieval misses. For an uncompressed memory, retrieval runs in the full embedding, and a two-dimensional map is the view in which a person audits what the agent stores, merges and deletes. The reversal test says which map answers which question, and NPCA adds axes named by the input variables.

\runhead{The user chooses the geometry, and the test answers for it.} In each of these uses, global structure is defined relative to a reference dissimilarity. Here the reference is the Euclidean or cosine distance for real data and the geodesic distance for the synthetic manifolds, where Isomap and the neighbor embeddings lead. A geodesic or diffusion distance on real data enters the profile in the same way. The profile is rank-based, so it is invariant to the stretched distances of t-SNE and UMAP.

\runhead{Conclusion.} For nearly two decades the choice of a map has rested on the claim that t-SNE and UMAP keep local structure and PCA global structure. Our strongest result is the LG reversal test, built on our DLP curve. It decides with a stated error rate which of two maps is local and which global, and where to change between them. It turns that claim into a result that users check on their own data. The test serves single-cell atlases, pathology, supply chains and the memories of AI agents, and other settings where a representation is compared with a reference. On twelve datasets of up to 1.38 million points the claim holds firmly on six, narrowly on three, within chance on two, and is reversed on one. On independent subsamples the test keeps its error rate at or below 0.020 (Figure~\ref{fig:sim}). We suggest reporting $(L,G)$ with published maps, and the reversal test whenever two maps are compared.

\section{Methods}\label{sec:methods}

\subsection{The locality profile, local and global fidelity}\label{sec:method}

One chance-corrected statistic, read at a small and at a large neighborhood size, gives a local and a global fidelity on the same scale.

\subsubsection{Setting and notation}

Let $\x_1,\dots,\x_n$ be data points with a reference dissimilarity $d_X$, the Euclidean or, for unit vectors, cosine distance in the chosen representation, and let $\y_1,\dots,\y_n\in\R^2$ be a map with Euclidean distance $d_Y$. For each point $i$ write $r^X_i(j)\in\{1,\dots,n-1\}$ for the rank of $j$ among the other points by increasing $d_X(\x_i,\cdot)$, and $r^Y_i(j)$ likewise in the map. For $1\le k\le n-2$ let $N^X_k(i)=\{j:r^X_i(j)\le k\}$ and $N^Y_k(i)$ be the $k$-nearest-neighbor sets. Superscripts $X$ and $Y$ mark the data and the map, superscripts $A$ and $B$ mark two methods, the subscript $k$ marks the neighborhood size, and the argument $(i)$ marks a single point. Supplementary Table~\ref{tab:notation} collects the notation.

At every scale the profile asks one question: what share of each point's $k$ nearest neighbors does the map keep, beyond chance?

\begin{definition}[Locality profile]
For point $i$ and scale $k$, let $O_k(i)=|N^X_k(i)\cap N^Y_k(i)|$ be the number of its $k$ nearest neighbors in the data that are also among its $k$ nearest in the map. The chance-corrected profile of point $i$ is
\begin{equation}\label{eq:nui}
\nu_k(i)=\frac{(n-1)\,O_k(i)/k-k}{n-1-k}.
\end{equation}
The locality profile of a map is the mean of these curves over points, $k\mapsto\nu_k$, $1\le k\le n-2$, indexed by the relative scale $s=k/n$. With the degree of locality preservation
\begin{equation}\label{eq:dlp}
\DLP_k=\frac1n\sum_{i=1}^n\frac{O_k(i)}{k},
\end{equation}
it reads
\begin{equation}\label{eq:nu}
\nu_k=\frac{(n-1)\DLP_k-k}{n-1-k}.
\end{equation}
\end{definition}

$\DLP_k$ over $k$ is the DLP curve \citep{han2022}, and $\nu_k$ is its chance-corrected form. From this curve we read $L$, $G$, the crossover, the test and $E$.

In plain terms, a map that scatters points at random scores 0 in expectation at every scale, a perfect map scores 1, and any monotone stretching of distances leaves the score unchanged.

\begin{proposition}[Chance level and invariance]\label{prop:chance}
Assume no ties among distances. (i) If the map carries no information about the data, so that each $N^Y_k(i)$ is a uniformly random $k$-subset of the other $n-1$ points, then $\mathbb{E}\,\DLP_k=k/(n-1)$ and $\mathbb{E}\,\nu_k=0$; a perfect map has $\nu_k=1$. (ii) $\nu_k$ is invariant under any rotation, reflection, translation, or uniform scaling of the map, and under any strictly increasing transformation of $d_X$ or $d_Y$.
\end{proposition}

\begin{proof}
(i) Fix $i$. Each of the $k$ members of $N^X_k(i)$ falls in the random set $N^Y_k(i)$ with probability $k/(n-1)$. By linearity of expectation, $\mathbb{E}|N^X_k(i)\cap N^Y_k(i)|=k^2/(n-1)$. Averaging over $i$ and dividing by $k$ gives $\mathbb{E}\,\DLP_k=k/(n-1)$. Equation~\eqref{eq:nu} is affine in $\DLP_k$ and vanishes at $\DLP_k=k/(n-1)$, so $\mathbb{E}\,\nu_k=0$. A perfect map has $N^Y_k(i)=N^X_k(i)$ for all $i$, so $\DLP_k=1$ and $\nu_k=\bigl((n-1)-k\bigr)/(n-1-k)=1$. (ii) A rotation, reflection or translation leaves every map distance unchanged, and a uniform scaling multiplies all of them by one positive constant. A strictly increasing transformation of $d_X$ or $d_Y$ keeps the order of the distances from each point. In every case the ranks $r^X_i(j)$ and $r^Y_i(j)$ are unchanged, and so are the sets $N^X_k(i)$ and $N^Y_k(i)$, $\DLP_k$ and $\nu_k$.
\end{proof}

The data bear this out. On 1,999 MNIST digits, 20 random maps give $\nu_k$ with mean $-0.0002$ and standard deviation 0.005 for $k$ from 10 to 999. Replacing each map distance $d_Y$ by $d_Y^3+e^{d_Y}$, a strictly increasing transformation, leaves $\nu_{10}$ and $\nu_{999}$ of PCA, t-SNE and SR-tSNE2 unchanged.

Invariance to monotone transformations matters for nonlinear embeddings, whose distances are distorted on purpose. t-SNE and UMAP expand the contrast between near and far pairs by an order of magnitude \citep{han2022}. This changes scores built on raw distances and leaves rank-based scores intact.

\runhead{From $\DLP_k$ to $L$ and $G$.} After the chance correction the profile has two readable ends, one local and one global. The raw curve first needs a correction for chance, and Figure~\ref{fig:concept}(b) shows why. $\DLP_k$ rises toward 1 for every map, because a neighborhood that holds most of the points shares most of them by chance; its chance level is $k/(n-1)$ (Proposition~\ref{prop:chance}). Equation~\eqref{eq:nu} subtracts this level and rescales, so that $\nu_k$ is 0 for a random map and 1 for a perfect one at every $k$ (Figure~\ref{fig:concept}c). At a small fixed neighborhood, $\nu_{10}$ says how many of each point's closest neighbors the map keeps beyond chance; this is $L$. At $k^\star=(n-1)/2$, Proposition~\ref{prop:duality} shows that $\nu_{k^\star}$ is the agreement on which half of the data is near each point. Proposition~\ref{prop:sheppard} ties this agreement to the rank correlation of distances, which global scores measure (Figure~\ref{fig:concept}d,e). $G$ averages the profile over $0.3\le k/n\le0.5$, a band that ends at this midpoint; the average keeps the meaning of the midpoint and lowers its variance.

\subsubsection{Near sets, far sets, and the upper end of the profile}

The upper end of the profile measures global structure. At the midpoint it is the agreement on which half of the data is near each point. This agreement is Blomqvist's rank correlation of the distances, and Kendall's $\tau$ under a Gaussian copula. The proof starts from a counting identity. At the midpoint, a map that keeps a point's near half has also kept its far half, so each pair asks one question, near or far.

\begin{proposition}[Near--far duality]\label{prop:duality}
Let $F^X_k(i)$ and $F^Y_k(i)$ be the complements of $N^X_k(i)$ and $N^Y_k(i)$ among the other $n-1$ points. Then $|F^X_k(i)\cap F^Y_k(i)|=n-1-2k+|N^X_k(i)\cap N^Y_k(i)|$. In particular, for $k^\star=(n-1)/2$ with $n$ odd,
\begin{equation}\label{eq:median}
\begin{aligned}
&\nu_{k^\star}=2h-1,\qquad h=\frac1n\sum_ih(i),\\
&h(i)=\frac{1}{n-1}\Bigl|\{j:\ \mathbf{1}[r^X_i(j)\le k^\star]=\mathbf{1}[r^Y_i(j)\le k^\star]\}\Bigr|,
\end{aligned}
\end{equation}
where $h(i)$ is the probability that the data and the map agree, for point $i$, on whether a random other point lies in the nearer or the farther half.
\end{proposition}

\begin{proof}
Among the other $n-1$ points, $|N^X_k(i)\cup N^Y_k(i)|=2k-|N^X_k(i)\cap N^Y_k(i)|$ by inclusion--exclusion. The set $F^X_k(i)\cap F^Y_k(i)$ is the complement of this union, which gives the first identity. Point $j$ is classified alike by the data and the map exactly when it lies in $N^X_k(i)\cap N^Y_k(i)$ or in $F^X_k(i)\cap F^Y_k(i)$. At $2k^\star=n-1$ the two sets have the same size, so the number of agreements is $2|N^X_{k^\star}(i)\cap N^Y_{k^\star}(i)|$ and $h(i)=|N^X_{k^\star}(i)\cap N^Y_{k^\star}(i)|/k^\star$. Averaging over $i$ gives $h=\DLP_{k^\star}$. With $n-1=2k^\star$, Equation~\eqref{eq:nu} reads $\nu_{k^\star}=(2k^\star\DLP_{k^\star}-k^\star)/k^\star=2\DLP_{k^\star}-1=2h-1$. For even $n$ the same count at $k=n/2$ gives $\nu_k=2h-1+O(1/n)$.
\end{proof}

On the same 1,999 digits, where $k^\star=999$, the identity holds to machine precision. $\nu_{k^\star}$ and $2h-1$ are both 0.328 for PCA and 0.183 for t-SNE and SR-tSNE2, and they differ by less than $10^{-15}$.

\runhead{Past the midpoint the profile mirrors itself.} The first identity of Proposition~\ref{prop:duality} also bounds the band of $G$. Let $\nu^F_j$ be the chance-corrected overlap of the $j$ farthest points, defined as in Equation~\eqref{eq:nu} with $F$ in place of $N$. Substituting the identity into Equation~\eqref{eq:nu} gives $\nu_k=\nu^F_{n-1-k}$ for every $k$. A neighborhood larger than half the data is the complement of a far set smaller than half, so above $k^\star$ the profile reads the far end of the same rankings again. On the 2,000 digits of Figure~\ref{fig:concept} the two sides differ by less than $3\times10^{-14}$ for PCA. $G$ therefore stops at the midpoint, where the near and far sets have the same size.

At small $k$ the profile asks whether a point keeps its nearest neighbors. At $k^\star$ it asks, for every point, which half of the data is near and which is far. The large distances decide this, so the answer depends on the layout of clusters, branches and regions. The next result connects this agreement to the rank correlation of distances, the usual measure of global structure. In short, a map that sorts every point's near and far halves correctly has medial correlation 1 with the data.

\begin{proposition}[Near--far agreement and rank correlation]\label{prop:sheppard}
Fix a point $i$ and let $(U_j,V_j)=\bigl(d_X(\x_i,\x_j),d_Y(\y_i,\y_j)\bigr)$ for the $n-1$ other points $j$. (i) $2h(i)-1$ is Blomqvist's medial correlation \citep{blomqvist1950} of these pairs: the share of pairs on the same side of both sample medians minus the share on opposite sides. (ii) If, given point $i$, the pairs are i.i.d.\ draws from a distribution with continuous margins and copula $\mathcal{C}_i$, then as $n\to\infty$
\begin{equation}\label{eq:blomqvist}
2h(i)-1\to\beta(i)=4\,\mathcal{C}_i\bigl(\tfrac12,\tfrac12\bigr)-1,
\end{equation}
a rank correlation of the two distances that needs no model. (iii) Under a Gaussian copula with correlation $\gamma(i)$,
\begin{equation}\label{eq:sheppard}
\beta(i)=\frac{2}{\pi}\arcsin\gamma(i),
\end{equation}
which is Kendall's $\tau$ of the pair and a strictly increasing function of its Spearman correlation $\rho_S(i)=\frac{6}{\pi}\arcsin\frac{\gamma(i)}{2}$.
\end{proposition}

\begin{proof}
(i) With $n-1=2k^\star$ and no ties, $r^X_i(j)\le k^\star$ holds exactly when $U_j$ lies below the sample median of the $U$'s, and likewise for $V$. By Proposition~\ref{prop:duality}, $h(i)$ is the share of pairs on the same side of both medians, and $2h(i)-1=h(i)-\bigl(1-h(i)\bigr)$ is the medial correlation. (ii) With continuous margins the sample medians converge to the population medians, and the share of pairs below both medians converges to $\mathcal{C}_i(\frac12,\frac12)$. The share above both medians equals $1-\frac12-\frac12+\mathcal{C}_i(\frac12,\frac12)=\mathcal{C}_i(\frac12,\frac12)$, so $h(i)\to2\mathcal{C}_i(\frac12,\frac12)$ and $2h(i)-1\to4\mathcal{C}_i(\frac12,\frac12)-1$ \citep{blomqvist1950}. (iii) For a Gaussian copula, Sheppard's formula \citep{sheppard1899} gives $\mathcal{C}_i(\frac12,\frac12)=\frac14+\frac{1}{2\pi}\arcsin\gamma(i)$, hence $\beta(i)=\frac{2}{\pi}\arcsin\gamma(i)$. For the bivariate normal, Kendall's $\tau$ equals $\frac{2}{\pi}\arcsin\gamma$ and Spearman's $\rho$ equals $\frac{6}{\pi}\arcsin\frac{\gamma}{2}$; both increase with $\gamma$.
\end{proof}

Proposition~\ref{prop:sheppard} explains why the upper end of the profile agrees with the rank correlation of distances, and the prediction holds on data. Over 400 MNIST points, $2h(i)-1$ tracks $\frac{2}{\pi}\arcsin\gamma(i)$ with Pearson correlation 0.94 (Figure~\ref{fig:concept}e). On real data $G$ ranks maps as the established global scores do (Results).

\subsubsection{Local fidelity, global fidelity, and derived quantities}

With the meaning of both ends established, the two scores follow. $L$ reads the profile at $k=10$ and $G$ averages it over $0.3\le k/n\le0.5$. Both lie on one scale, so they can be compared directly.

\begin{definition}[Local and global fidelity]
For a map of $n$ points, the local fidelity and the global fidelity are
\begin{equation}\label{eq:LG}
L=\nu_{10},\qquad G=\frac{1}{|\mathcal K|}\sum_{k\in\mathcal K}\nu_k,\quad \mathcal K=\{k:\ 0.3\le k/n\le0.5\}.
\end{equation}
\end{definition}

\runhead{The chance correction reverses the order of the maps.} On the 70,000 MNIST digits the chance correction reverses the order of the maps between the two ends (Supplementary Fig.~\ref{fig:LGcalc}). Every $\DLP_k$ curve ends at 1, and near $k=n-1$ it is close to its chance level, so the raw curve hides the differences at the upper end (panel a). The chance correction removes the level $k/(n-1)$ and spreads the profiles apart (panel b). At $k=10$, $L$ is 0.390 for SR-tSNE2 and 0.377 for t-SNE, against 0.007 for PCA and 0.008 for NPCA. Over the band, the order reverses: PCA has $G=0.301$ and NPCA 0.266, against 0.187 for SR-tSNE2 and for t-SNE (panel c).

\runhead{$L$ and $G$ are compared directly.} Because $L$ and $G$ are the same statistic at two scales, they are set side by side for one map and compared between maps of the same data. $L$ uses an absolute neighborhood and $G$ a band of relative scale, so comparisons are made within a dataset, and the reversal test decides them for two maps. $L$ is read at a fixed small neighborhood, the scale at which t-SNE, UMAP, and their relatives are designed to be faithful. $G$ averages the profile over its upper end, where Proposition~\ref{prop:duality} makes it a statement about near and far halves. Two methods can then be compared on both axes at once.

\runhead{A relative scale makes $G$ estimable from a subsample.} $G$ is defined at a scale relative to $n$, which makes it comparable across datasets and estimable from a subsample. A fixed $k$ is a global scale on a small dataset and a local one on a large dataset; $k=50$ is a third of 150 points and 0.005 percent of a million. Take a relative scale $s=k/n$ and a uniform subsample of $m$ points. The $\lfloor sm\rfloor$ nearest points within the subsample approximate the $s$-quantile neighborhood in the full data. $G$ computed on the subsample therefore estimates $G$ of the full map. $L$ at fixed $k=10$ behaves differently: on the subsample it refers to the relative scale $10/m$, which grows as $m$ shrinks. Both are measured below.

\runhead{$E$ reads the profile at each point's entity size.} Many tasks ask for a whole entity: all cells of a type, all records of a customer, all passages on a topic. The scale of such a task lies between $L$ and $G$, and it changes from entity to entity. We therefore read the profile at each point's own entity size.

\begin{definition}[Entity fidelity]
Let point $i$ belong to entity $c(i)$ with $n_{c(i)}$ members, and let $k_i=n_{c(i)}-1$, with $1\le k_i\le n-2$. With $o_i=|N^X_{k_i}(i)\cap N^Y_{k_i}(i)|/k_i$, the entity fidelity of point $i$ and of the map are
\begin{equation}\label{eq:E}
E(i)=\frac{(n-1)\,o_i-k_i}{n-1-k_i},\qquad E=\frac1n\sum_{i=1}^nE(i).
\end{equation}
\end{definition}

$E$ applies the chance correction of Equation~\eqref{eq:nu} at a scale that varies with the point, so it is corrected point by point before the average. $E$ uses the entity sizes alone. When every entity has $n_0$ members, $E=\nu_{n_0-1}$, a single point of the profile; $E$ adds most where entity sizes differ.

\runhead{Two quantities compare maps.} Two derived quantities summarize comparisons. The crossover scale says where the other map becomes the more faithful one. The exchange rate says how much global fidelity a parameter gains for each unit of local fidelity it trades. Both need the two ends on one scale.

\begin{definition}[Crossover scale and exchange rate]
The crossover scale of a local method $A$ against a global method $B$, with profiles $\nu^A_k$ and $\nu^B_k$, is
\begin{equation}\label{eq:crossover}
s^\times(A,B)=\min\{k/n:\ \nu^B_k\ge\nu^A_k\}.
\end{equation}
If $\nu^A_k>\nu^B_k$ at every $k$, there is no crossover and $A$ is the one map to use.
A parameter path is the sequence of maps of one method as one parameter $\theta$ changes. The exchange rate of a path from $\theta_0$ to $\theta_1$, with fidelities $(L_0,G_0)$ and $(L_1,G_1)$ and $L_1\ne L_0$, is
\begin{equation}\label{eq:rate}
\eta=-\frac{G_1-G_0}{L_1-L_0}.
\end{equation}
\end{definition}

Below $s^\times$ method $A$ is the more faithful map, and above it $B$ is. The reversal test turns this reading into a test. The rate $\eta$ is the global fidelity gained per unit of local fidelity traded, for example when t-SNE's perplexity rises from 30 to 200. The rate is defined because $L$ and $G$ are the same statistic, so their differences share one unit.

\subsubsection{Computation}

All these quantities come from one profile. The whole profile follows from one cumulative sum, and on large data $G$ is estimated from a subsample. Point $i$ keeps neighbor $j$ at every $k\ge \mu_i(j)=\max\{r^X_i(j),r^Y_i(j)\}$, so its overlap $|N^X_k(i)\cap N^Y_k(i)|$ is the number of $j$ with $\mu_i(j)\le k$. One histogram of $\mu_i(j)$ per point and its cumulative sum give the whole profile of every point, and their mean gives $\nu_k$ at every $k$. This needs the two $n\times n$ rank matrices, $O(n^2\log n)$ time and $O(n^2)$ memory; for $n=5{,}000$ this takes a few seconds on one core. For larger $n$, $G$ is estimated on a uniform subsample of a few thousand points, which the next subsection shows to be accurate. The code is released with the paper.

\subsubsection{$G$ is estimated from a few thousand points}\label{sec:subsample}

Because $G$ is defined on a relative scale, a uniform subsample of a few thousand points estimates it at any $n$, with fixed time and memory. On each large dataset we drew five uniform subsamples of each size $m=1{,}000$, 2,000, 4,000 and 8,000 points for the PCA and t-SNE maps. $G$ was computed on each subsample from the subsampled data and map alone (Supplementary Fig.~\ref{fig:robust}c,d).

At $m=2{,}000$ the mean estimate lies within 0.010 of its value at $m=8{,}000$ on every dataset and for both maps, and at $m=4{,}000$ within 0.005. The standard deviation over draws is at most 0.012 at $m=2{,}000$. On SEA-AD, with 1.38 million nuclei, $G$ of PCA is 0.287, 0.288, 0.285 and 0.286 at the four sizes, and that of t-SNE 0.241, 0.240, 0.237 and 0.242 (Supplementary Fig.~\ref{fig:robust}d). We therefore report $G$ as the mean over ten subsamples of 2,000 points, which takes seconds for any $n$. $L$ is defined on an absolute scale. At fixed $k=10$, a subsample of 2,000 points asks about the nearest 0.5 percent of the data. On a million points the same $k$ asks about the nearest 0.001 percent, a far more local scale. $L$ is therefore computed exactly from the 10 nearest neighbors of every point.

\subsection{How the test works}\label{sec:testworks}

The test has three steps. Per-point scores turn the comparison into paired means, two one-sided tests give the declaration, and a block jackknife gives standard errors for dependent points.

\runhead{Step 1: every score is a mean over points.} From the profile of each point (Equation~\eqref{eq:nui}) define
\begin{equation}\label{eq:LGi}
L(i)=\nu_{10}(i),\qquad G(i)=\frac{1}{|\mathcal K|}\sum_{k\in\mathcal K}\nu_k(i).
\end{equation}
The score of a whole map is then the average of the scores of its points.

By Equations~\eqref{eq:nui} and~\eqref{eq:nu}, averaging $\nu_k(i)$ over points gives $\bigl((n-1)\DLP_k-k\bigr)/(n-1-k)=\nu_k$, so
\begin{equation}\label{eq:pp}
\nu_k=\frac1n\sum_i\nu_k(i),\qquad L=\frac1n\sum_iL(i),\qquad G=\frac1n\sum_iG(i).
\end{equation}

On the 1,999 MNIST digits, the mean of $\nu_{10}(i)$ over points equals $\nu_{10}$ to machine precision for PCA, t-SNE and SR-tSNE2. Equation~\eqref{eq:pp} is what makes standard errors possible: on MS MARCO the per-point differences give a standard error of 0.009 for the local difference and 0.012 for the global one (Results).

By Equation~\eqref{eq:pp}, comparing two maps of the same data means comparing paired means. For methods $A$ and $B$, the differences $d^L(i)=L^A(i)-L^B(i)$ and $d^G(i)=G^B(i)-G^A(i)$ have means $L^A-L^B$ and $G^B-G^A$.

\runhead{Step 2: a reversal needs both one-sided tests to reject.} A reversal is declared only when the data support both claims at once.
\begin{definition}[LG reversal test]\label{def:lgtest}
For methods $A$ and $B$ applied to the same data, the local test has null hypothesis $H^L_0:\ L^A\le L^B$ and the global test has null hypothesis $H^G_0:\ G^B\le G^A$. $A$ is declared local and $B$ global, a reversal, when both nulls are rejected at level $\alpha$. With $p_L$ and $p_G$ the p-values of the two tests, the p-value of the declaration is $p=\max(p_L,p_G)$.
\end{definition}

By the intersection--union principle \citep{berger1982iut}, the declaration has level $\alpha$: a false declaration needs a false rejection, and each test makes one at rate at most $\alpha$. The two tests therefore run at the full level, with no correction.

The level holds on data: over 48 one-sided global tests on the large datasets the rejection rate of a true null averages 0.046. On 50 independent subsamples the local test rejects at no more than 0.020 (Calibration, below). A declaration requires both: $A$ more local and $B$ more global. On the Swiss roll, for example, t-SNE stays above PCA at every scale, so t-SNE is the one map to use.

\runhead{Step 3: a block jackknife allows for shared neighbors.} Two points that are near each other share neighbors, so a map that tears one region apart lowers the $L(i)$ of every point in it, and the per-point differences are dependent. Write $d(i)$ for either $d^L(i)$ or $d^G(i)$ and $\bar d$ for its mean over all $n$ points. We cut the data into $C=20$ blocks by $k$-means in the input space, so that most shared neighborhoods fall inside one block. For the upper end, the calibration below checks the resulting standard errors against a null that resamples the same blocks. The delete-a-group jackknife \citep{efron1982jackknife} leaves out one block $c$ at a time and recomputes the mean difference $\bar d_{(c)}$ from the remaining points. With $\bar d_{(\cdot)}$ the mean of the $C$ values $\bar d_{(c)}$, it estimates
\begin{equation}\label{eq:jk}
\widehat{\mathrm{SE}}^2=\frac{C-1}{C}\sum_{c=1}^{C}\big(\bar d_{(c)}-\bar d_{(\cdot)}\big)^2 .
\end{equation}
The statistic $t=\bar d/\widehat{\mathrm{SE}}$ is compared with a $t$ distribution with $C-1$ degrees of freedom. Its upper-tail probability is the one-sided p-value: $p_L$ for the local test and $p_G$ for the global test. The neighbor sets are computed once from all $n$ points, and the jackknife leaves out blocks of the points over which the per-point scores are averaged.

\runhead{The crossover interval.} The same leave-one-block-out means give an interval for the crossover. With $\bar\nu^A_{(c)}$ and $\bar\nu^B_{(c)}$ the profiles averaged over the points outside block $c$, $s^\times_{(c)}$ is read from them by Equation~\eqref{eq:crossover}. The interval is $s^\times\pm t_{C-1,0.975}\,\widehat{\mathrm{SE}}(s^\times)$, where $t_{C-1,0.975}$ is the 0.975 quantile of the $t$ distribution with $C-1$ degrees of freedom and $\widehat{\mathrm{SE}}(s^\times)$ is Equation~\eqref{eq:jk} applied to the values $s^\times_{(c)}$. When the profile of $A$ stays above that of $B$ at every scale, the test reports $A$ as the one map to use.

\subsubsection{Calibration}

The global test rejects a true null at close to its stated rate. We built a null with the dependence of each dataset. The per-point differences $d(i)$ were centered so that their mean is exactly zero. The 20 blocks were then resampled with replacement, 500 times on the large datasets and 2,000 times on the small ones, and the jackknife statistic was recomputed each time. On the large datasets the test runs on a uniform subsample of 5,000 points. On this subsample $k=10$ is a coarser relative scale than in the exact $L$ of Supplementary Table~\ref{tab:largeLG}, so the local differences of the test are read at that scale. On MS MARCO the local difference of t-SNE is 0.213 in the test and 0.253 in the table. Over the 48 one-sided global tests (twelve datasets, four maps against PCA) the rejection rate at level 0.05 averages 0.046. On the three smaller datasets and the three manifolds, tested on all points, the rejection rate of the 36 one-sided tests averages 0.051. On 50 independent subsamples of each large dataset, with the maps fixed, the local test rejects a true null at no more than 0.020 and the global test at 0. A test that treats points as independent rejects at up to 0.20 and 0.48 (Figure~\ref{fig:sim}, Supplementary Table~\ref{tab:sim} and Supplementary Note~\ref{app:localcal}). The two kinds of error answer different questions. The spread over subsamples is the error of drawing other points from the same dataset. The block jackknife also counts the variation between regions of the data, the error of drawing another dataset from the distribution the data come from, and that distribution is what the test makes inference about. Its standard errors therefore exceed the spread over subsamples, by a median factor of 4.3 for $L$ and 6.3 for $G$, by design. Supplementary Note~\ref{app:blocks} repeats the declarations with 10 and 50 blocks. Refitting t-SNE on MNIST with five more seeds moves the global difference by a standard deviation of 0.010, about a sixth of its jackknife standard error. Four of the six maps are declared, and the other two reach $p\le0.061$ (Supplementary Note~\ref{app:refit}). The test makes inference for the two fitted maps.

\subsection{Experimental design}\label{sec:setup}

The main claims of this paper are tested on 156 maps of twelve datasets of 50,000 to 1.38 million points: seven methods on each dataset and 72 maps from 24 parameter settings on MNIST.

\runhead{Twelve large datasets, four of them applications.} The study rests on twelve datasets of 50,000 to 1.38 million points (Supplementary Table~\ref{tab:largedata}). Four are applications, which lead every table and figure, and eight are the benchmark datasets of our earlier SR-tSNE study \citep{han2026srtsne}. The four applications are as follows. For pathology, Pathology-100K holds 100,000 colorectal tissue tiles (NCT-CRC-HE-100K)\citep{kather2019} embedded with Phikon \citep{filiot2023phikon}. For single cells, SEA-AD holds the 1,378,211 nuclei of the SEA-AD middle temporal gyrus atlas \citep{gabitto2024seaad}, as 50 principal components. For supply chains, Instacart holds the purchases of 195,827 Instacart customers over the 2,000 products bought by the most customers. Each customer has at least five of these products and is a unit vector of $\log(1+\text{count})$. For AI agent memory, MS MARCO holds one million passages drawn at random from the 8.8 million of MS MARCO \citep{bajaj2016msmarco}. They are embedded with BGE-base \citep{xiao2024cpack} as the long-term memory of a retrieval agent. Among the benchmarks, MNIST \citep{lecun1998gradient} and Fashion-MNIST \citep{xiao2017fashion} enter as pixels and Covertype \citep{blackard1999comparative} as its 54 attributes. CIFAR-10 \citep{krizhevsky2009learning} enters as 512 pooled features of an ImageNet-trained ResNet-18 \citep{he2016deep}, Speech Commands \citep{warden2018speech} as mean-pooled wav2vec~2.0 features \citep{baevski2020wav2vec}, and AG News \citep{zhang2015character} as mean-pooled BERT embeddings \citep{devlin2019bert}. Renal-Cortex and Brain-Cortex are the two single-cell datasets of the SR-tSNE study, each reduced to 50 principal components of the log-normalized expression of 2,000 variable genes. The reference dissimilarity is the Euclidean distance in the representation of Supplementary Table~\ref{tab:largedata}, and the cosine distance for unit vectors.

\runhead{Smaller datasets and manifolds with known structure.} Two further groups of data complete the study. Three smaller datasets of a few thousand points show the scores at the size an analyst often works with (Results). Pathology-7K holds 7,180 colorectal tissue tiles in nine classes (CRC-VAL-HE-7K)\citep{kather2019}, embedded with Phikon. Online Retail holds the purchases of 5,459 customers of a UK online wholesaler over 2,000 products, from about one million invoice lines \citep{chen2012retail}. Each customer is a unit vector of $\log(1+\text{quantity})$. 8-K holds 5,000 SEC 8-K filings, drawn at random from 18,329 and embedded with BGE-base. Three synthetic manifolds have a known global structure. The Swiss roll and the S-curve have 1,500 points each, and their reference is the Euclidean distance between the unrolled two-dimensional coordinates, the true geodesic distance. The tree is a ten-branch diffusion-limited aggregation in 100 dimensions with 2,000 points, generated with the PHATE package. Its reference is the shortest-path distance in the 10-nearest-neighbor graph. On the Swiss roll and the S-curve the true global structure is the unrolled sheet.

\runhead{Seven maps on every large dataset.} Seven maps are fitted on every large dataset, on all points with a fixed seed. The first is PCA. Nonnegative PCA (NPCA) \citep{han2010tcbb}, our earlier method, maximizes variance under nonnegative loadings by projected gradient ascent (two components, orthonormality penalty 0.1, 3,000 steps). t-SNE uses openTSNE \citep{policar2024}, accelerated by the fast Fourier transform, with PCA initialization, which shapes its global layout \citep{kobak2021}. It runs at perplexity 30, the default, and at perplexity 200 (t-SNE$_{200}$). SR-tSNE2 \citep{han2026srtsne}, our rank refinement of t-SNE, refines the perplexity-30 t-SNE map in $O(n\log n)$ time. It adds a smooth rank loss on each point's 10 nearest neighbors in the input (learning rate 50, rank weight $\lambda=20$, 2,000 iterations). UMAP and PaCMAP run at their package defaults. On the 60,000 MNIST training digits, four parameter paths vary one setting each. They are t-SNE's perplexity (5 to 500), UMAP's number of neighbors (5 to 500), the exaggeration of t-SNE (1 to 32), and the number of principal components before t-SNE (2 to 50). Each setting has three runs, one with PCA initialization and two random, which gives 72 further maps. The three smaller datasets are embedded by the same methods except t-SNE$_{200}$, and also by PHATE and DREAMS \citep{kury2026}. On them SR-tSNE2 runs at the setting of the SR-tSNE study for fewer than 50,000 points (rank weight $\lambda=5$, 1,000 iterations). The three manifolds are embedded by PCA, t-SNE, SR-tSNE2, UMAP and PaCMAP, and also by kernel PCA (KPCA), LLE \citep{roweis2000}, Isomap \citep{tenenbaum2000}, PHATE and TriMAP \citep{amid2019}.

\runhead{$L$ is exact and $G$ is averaged over ten subsamples.} $L$ is computed exactly over all points from the exact 10 nearest neighbors in the data, found on a GPU. $G$ and the profile $\nu_k$ are the means over 10 uniform subsamples of 2,000 points, which the subsample check above shows is enough. From the profile we also read the average over the middle band $0.1\le k/n\le0.2$, the midpoint value and two alternative global bands ($0.2$--$0.5$ and $0.4$--$0.5$). We also compute the area under $R_{NX}$ on a logarithmic axis, the usual summary of neighbor curves \citep{lee2015}. As established global scores we compute two quantities on the same subsamples. The distance rank correlation (distance $\rho$ in the figures) is the Spearman correlation $\rho$ between pairwise map distances and the reference dissimilarities. Random-triplet accuracy is the fraction of random triplets $(i,j,l)$ for which data and map agree on whether $j$ or $l$ is closer to $i$ \citep{wang2021}. O-DLP@10 is the share of each point's 10 nearest neighbors that the map returns at their exact rank, averaged over points.

\runhead{The task retrieves each entity from its centroid.} For entity $c$, the exact selection is the $n_c$ points nearest to the centroid of $c$ in the data. The map selection is the $n_c$ points nearest to its centroid on the map. The miss rate is the share of the exact selection that the map does not return. For each pair of an entity and a map we average $E(i)$, $L$, $G$ and a common-scale reading $M$ over up to 200 sampled members of the entity (seed 0). All four come from the exact rankings of the $n-1$ other points. $M$ reads the profile at one $k$ for every point, the median of $n_{c(i)}-1$ over all points. $M$ separates the effect of reading each point at its own entity size. Over the seven maps of each dataset we report the partial rank correlation of the miss rate with $E$ given $L$, $G$ and $M$. Its 95\% interval comes from a bootstrap over entities (2,000 resamples).

{\small \bmhead{Supplementary information} Supplementary Notes, Supplementary Results, Supplementary Figs.~1--11 and Supplementary Tables~1--12 follow the references.
\bmhead{Data availability} All data are public. Pathology-100K and Pathology-7K are NCT-CRC-HE-100K and CRC-VAL-HE-7K \citep{kather2019}. SEA-AD is the middle temporal gyrus atlas of the Seattle Alzheimer's Disease Brain Cell Atlas \citep{gabitto2024seaad}. The endothelial cells come from the Tabula Sapiens atlas \citep{tabulasapiens2022}, as released in the CZ CELLxGENE Census. Instacart is the Instacart Online Grocery Shopping Dataset 2017, released by Instacart and distributed on Kaggle. Online Retail is Online Retail II of the UCI Machine Learning Repository \citep{chen2012retail}. MS MARCO \citep{bajaj2016msmarco}, FinQA \citep{chen2021finqa}, LoCoMo \citep{maharana2024locomo}, MNIST, Fashion-MNIST, CIFAR-10, Speech Commands, Covertype and AG News come from their public releases (Methods). The 8-K filings come from SEC EDGAR. Renal-Cortex and Brain-Cortex are the single-cell benchmarks of the SR-tSNE study, which gives their sources \citep{han2026srtsne}.
\bmhead{Code availability} The package \texttt{lgfid} computes $L$, $G$, O-DLP@10 and the profile of any map from the data and the map. It runs the reversal test (Methods) on two maps, from a command line or from Python. It computes $L$ exactly and $G$ from ten subsamples of 2,000 points, as in Methods, and its tests check the identity map and a random map. Data and code: \url{https://github.com/hank08819/LG2Test}.

\bmhead{Author contributions} H.H. designed the study, ran the experiments and wrote the paper.
\bmhead{Competing interests} The author declares no competing interests.
}
\bibliographystyle{sn-nature}
\bibliography{refs}

\clearpage
\let\FloatBarrier\origFloatBarrier
\setcounter{secnumdepth}{3}
\setcounter{section}{0}\setcounter{figure}{0}\setcounter{table}{0}\setcounter{equation}{0}
\titleformat{\section}{\sffamily\bfseries\fontsize{11}{13}\selectfont}{\thesection}{0.6em}{}
\titleformat{\subsection}{\sffamily\bfseries\fontsize{9}{11}\selectfont}{\thesubsection}{0.6em}{}
\titleformat{\subsubsection}{\sffamily\bfseries\itshape\fontsize{8.5}{10}\selectfont}{\thesubsubsection}{0.6em}{}
\captionsetup[figure]{name=Supplementary Fig.,labelsep=pipe}\captionsetup[table]{name=Supplementary Table,labelsep=pipe,position=top}
\makeatletter\renewenvironment{table}[1][tp]{\@dblfloat{table}[tp]}{\end@dblfloat}\makeatother
\twocolumn[{%
\noindent{\sffamily\footnotesize\bfseries Supplementary information}\par\vspace{5mm}
\noindent{\sffamily\bfseries\fontsize{17.5}{21}\selectfont \mbox{Quantifying local and global fidelity of nonlinear embeddings}\par}\vspace{4mm}
\noindent{\sffamily\fontsize{10}{12}\selectfont Henry Han}\par\vspace{1mm}
\noindent{\sffamily\scriptsize Data Science and AI Innovation Laboratory, Baylor University, Waco, Texas, USA.\quad e-mail: Henry\_Han@baylor.edu}\par\vspace{6mm}
}]
\section{Supplementary notes}\label{app:notes}

\subsection{Related work}\label{sec:related}

Our DLP curve \citep{han2022} is the curve from which every score here is read. Earlier work supplies the global scores we use to validate $G$, and other neighbor-based quality curves, reviewed below. Three things are new here, and all three come from our DLP curve: a proven meaning for its upper end, local and global scores on one scale, and an error rate for comparing two maps.

\runhead{Rank-based quality curves.} \citet{lee2009} unified several neighborhood-based quality criteria through the co-ranking matrix and defined $Q_{NX}(k)$, the average overlap of $k$-neighborhoods. Its chance-corrected rescaling, $R_{NX}(k)$, appears in later work of the same group. \citet{lee2015} summarized $R_{NX}(k)$ by its area on a $\log k$ axis to compare multiscale neighbor embeddings. Trustworthiness and continuity \citep{venna2001} penalize intrusions and extrusions by rank and are usually reported at a single small $k$. NeRV \citep{venna2010} casts a map as an information-retrieval display and trades the precision of retrieved neighbors against their recall. The coRanking package \citep{kraemer2018} computes these co-ranking criteria in R, and two surveys compare projection techniques with many quality measures \citep{espadoto2021,nonato2019}. In single-cell genomics, \citet{heiser2020} scored dimension reductions by how well they preserve the structure of the data, and \citet{chari2023} showed that reducing such data to two dimensions inevitably distorts it, neighborhoods included. Our DLP curve \citep{han2022} is the starting point of this paper, because one statistic covers every scale. Their summaries weight the lower end: the area under $R_{NX}$ is dominated by small $k$ by design. \citet{lee2009} split $Q_{NX}$ at a data-dependent size $k_{\max}$ into $Q_{\text{local}}$ and $Q_{\text{global}}$ and advised consulting $Q_{\text{global}}$ to break ties. \citet{lueks2011} found that this global measure depends on absolute ranks and one parameter, and replaced it with a weighting governed by two user-set thresholds. Our $G$ keeps the single curve, reads it on a fixed band of relative scale after the chance correction, and gives its upper end a proven meaning.

\runhead{Global structure of neighbor embeddings.} \citet{kobak2019} and \citet{kobak2021} showed that the global layout of t-SNE and UMAP maps is largely set by initialization, and that large perplexities or multiscale affinities improve it. They measured global structure by the distance rank correlation, often between cluster centroids. \citet{bohm2022} placed t-SNE, UMAP and related methods on a spectrum of attraction against repulsion, which the exaggeration of t-SNE traverses. \citet{wang2021} analyzed the losses of t-SNE, UMAP and TriMAP \citep{amid2019}, introduced PaCMAP, and measured global structure by random-triplet accuracy. \citet{huang2022} compared methods for transcriptomic data on separate local and global criteria. These studies shaped what is now believed about each method. Their evidence pairs a local score with a global score of a different kind; weighing a gain at one end against a change at the other calls for one scale. DREAMS \citep{kury2026} comes closest: it traces a path between t-SNE and PCA in the plane of neighbor recall and distance rank correlation. Its two axes keep their own ranges and chance levels. Our $(L,G)$ plane puts both axes on one scale, which defines the crossover and the exchange rate, and the Results check $G$ against the global scores these studies used.

\runhead{Compressed memories of AI agents.} Compressed embeddings, from truncated \citep{kusupati2022matryoshka} to product-quantized \citep{jegou2011pq}, are judged by the recall of their nearest neighbors \citep{aumuller2020annbench}. Recall is a local score. Agents also consolidate memories and answer questions from their organization into topics \citep{edge2024graphrag}, and these tasks depend on global structure. With the uncompressed embedding as the reference, the profile scores a compressed memory at both ends.

\runhead{Comparing two maps.} The studies above compare maps by point values of their scores. Our LG reversal test adds inference to this comparison: a standard error for the difference between two maps of the same data, and a decision on whether one map is local and the other global. Neighboring points share neighbors; a block jackknife \citep{efron1982jackknife} handles this dependence, and the intersection--union principle \citep{berger1982iut} combines the two one-sided tests.

The Methods define $L$ and $G$ on one chance-corrected profile, prove what its upper end measures, and build the LG reversal test on the per-point form of the same scores.

\subsection{Contributions}\label{app:contrib}
\runhead{Contributions.} We apply the scores and the test to twelve datasets of 50,000 to 1.38 million points. Four come from applications: Pathology-100K, 100,000 pathology tiles; SEA-AD, 1.38 million single nuclei; Instacart, the purchases of 195,827 customers of a grocery supply chain; and MS MARCO, the one million passages above. The other eight are benchmarks of images, speech, text and single cells. Three smaller datasets complete the study, among them the Online Retail purchases. So do three synthetic manifolds, among them the Swiss roll and the S-curve. The contributions are listed by importance, and each rests on the one before it.
\begin{enumerate}[leftmargin=1.5em,itemsep=1pt]
\item \emph{A new test of whether one map is local and another global.} The LG reversal test turns the usual claim, that t-SNE is local and PCA global, into a decision with a stated error rate. Its global test holds its level, with a mean rejection rate of 0.046 over 48 tests at level 0.05. It also gives the crossover size, with a 95\% interval, at which a user should change maps. On the memory of a retrieval agent it gives that size in passages (Results, Figure~\ref{fig:agentmem}b--d).
\item \emph{New definitions of local and global fidelity on one scale, with a proven upper end.} The test is possible because $L$ and $G$ are one chance-corrected statistic at two scales. We prove that at $k=(n-1)/2$ the profile is the agreement on which half of the data is near each point. It converges to a rank correlation of the distances for i.i.d.\ pairs with continuous margins, and to Kendall's $\tau$ under a Gaussian copula (Propositions~\ref{prop:duality} and~\ref{prop:sheppard}). The common scale defines two new quantities, the crossover scale and the exchange rate (Methods).
\item \emph{Validation on 156 maps of up to 1.38 million points.} The propositions lead us to expect that $G$ ranks maps as the established global scores do, and the data agree. $G$ has rank correlation 0.89 with the distance rank correlation and 0.96 with random-triplet accuracy. Over 84 maps the area under $R_{NX}$ reaches 0.22 and 0.30, against 0.84 and 0.93 for $G$. The value of $G$ changes little with the band, and 2,000 points estimate it to within 0.01.
\item \emph{Tests of the usual claims in four applications.} With a validated $G$ and the test, the usual claims can be checked. The test declares t-SNE local and PCA global on seven of the twelve datasets, with the crossover at 9 to 20 percent of the data. On Instacart, Online Retail, the Swiss roll and the S-curve the test finds the neighbor embeddings more global. PCA has the highest $G$ on six datasets and PaCMAP on four. The parameters said to add global structure have measured exchange rates and estimated energy costs. On the 60,000 MNIST digits PCA keeps the highest $G$ and t-SNE the highest $L$ across all 24 settings (Results and Supplementary Note~\ref{sec:findings}).
\item \emph{Our own methods on the same scale.} The scale also measures design choices. SR-tSNE2 \citep{han2026srtsne}, an $O(n\log n)$ rank refinement of t-SNE, has the highest $L$ on all twelve datasets. It returns neighbors at their exact rank 18\% to 76\% more often than t-SNE, with $G$ within 0.0005. Nonnegative PCA \citep{han2010tcbb} names its axes by tissue, cell type or product, with $G$ within 0.14 of PCA's (Supplementary Note~\ref{sec:findings}).
\item \emph{Software.} The Python package \texttt{lgfid} returns $L$, $G$, the profile and the reversal test with its crossover interval for any data and one or two maps. It computes them exactly up to 5,000 points and from subsamples beyond.
\end{enumerate}

\section{Supplementary results}\label{app:results}

\subsection{The LG reversal test: which map answers which question}

\runhead{The question every user faces.} Analysts often keep two maps of the same data and use each for a different purpose. A pathologist searches one map for tiles that resemble a tumor tile and reads the layout of tissues from another. The keeper of an AI agent's memory inspects the passages near a query on one map and the grouping of topics on another. A supply-chain planner finds customers who buy like a given customer, and also segments all customers for stocking. Each of them must decide which map answers which question (Supplementary Fig.~\ref{fig:cartoon}).

\runhead{The test puts evidence behind the choice.} The choice is usually made by citing the design of a method or by looking at a map. The test adds an error rate. Two profiles computed on a finite sample can cross by chance, through the choice of points drawn. On SEA-AD, PCA is more global than t-SNE by 0.044 in $G$, a gap that looks decisive on a plot, and the test places it within chance ($p=0.16$).

\runhead{The test is new.} Earlier quality curves and scores \citep{venna2001,lee2009,lee2015,wang2021} report point values for one map at a time. The LG reversal test adds what a comparison of two maps needs: a decision on whether they reverse their order between the two ends, and an interval for the scale at which they cross. The test is possible here because $L$ and $G$ are the same statistic at two scales, and both are means of per-point scores.

\subsubsection{What the test returns}

\runhead{The four applications.} Results applies the test to every dataset, and the four applications show what it gives a user.
\begin{itemize}
\item \emph{Pathology.} On Pathology-100K t-SNE is local and PCA global ($p=1.4\times10^{-7}$), with the crossover at 10,040 tiles (4,210 to 15,870). The search for tiles that resemble a tumor tile uses the t-SNE map, and the layout of the nine tissues uses the PCA map.
\item \emph{AI agent memory.} On MS MARCO t-SNE is local and PCA global ($p=2.6\times10^{-5}$), with the crossover near 150,000 passages (Results). The passages near a query are inspected on the t-SNE map, and the layout of topics for consolidation on the PCA map (Figure~\ref{fig:agentmem}b--d).
\item \emph{Supply chains.} On Instacart t-SNE is more local than PCA by 0.080 and also more global by 0.021, so t-SNE leads at both ends and no reversal is declared ($p=0.97$).
\item \emph{Single cells.} On SEA-AD the global gap between PCA and t-SNE stays within its sampling error ($p=0.16$). t-SNE serves the neighbors of a cell, and PaCMAP, the map with the highest $G$, the distances between cell types.
\end{itemize}

\runhead{Protection against biased conclusions.} The test guards against three sources of bias in common practice. The first is to take the design of a method as a property of the data; on Instacart, Online Retail, the Swiss roll and the S-curve the test finds the neighbor embeddings more global. The second is to read a chance gap as a difference; the test places the SEA-AD gap within chance. The third is to treat neighboring points as independent; a block jackknife gives standard errors of the right size. The global test holds its level: over 48 global tests the rejection rate under the null averages 0.046 at level 0.05.

\subsection{The upper end of the profile measures global structure}

\runhead{Two profile shapes.} Two shapes recur in the profiles of the seven methods (Supplementary Fig.~\ref{fig:profiles}). The profile of PCA rises on every dataset, from its lowest value at the nearest neighbors to its highest at the near and far halves. The profile of t-SNE is higher at its lower end than in the global band on five of the twelve. Local and global behavior thus appear as the two ends of one profile, and the question is whether its upper end measures what the established global scores measure.

\subsubsection{$G$ is stable across bands and subsamples}\label{sec:robust}

A useful global score also stays put when its band, its random seed and its sample change. This subsection checks the band and the seed, and the next one the sample.

\runhead{The conclusions hold for every global band tested.} Other bands give nearly the same scores as $0.3\le k/n\le0.5$. Over the 84 maps of the large datasets, $G$ correlates at 0.98 with the $0.2$--$0.5$ band and with the $0.4$--$0.5$ band, and at 0.96 with the single midpoint value (Supplementary Fig.~\ref{fig:robust}a). Narrower bands at the top correlate more closely with the distance rank correlation, 0.91 for $0.4$--$0.5$ and 0.94 for the midpoint, and the $0.3$--$0.5$ band averages over more scales. Within a dataset, the global bands rank the methods as $G$ does, with mean Kendall $\tau$ of 0.83 and 0.89. The middle band $0.1\le k/n\le0.2$ gives a different ranking ($\tau=0.15$), so the ranking is set at the upper end. Supplementary Table~\ref{tab:sens} in Supplementary Note~\ref{app:sensitivity} gives the full comparison.

\runhead{$G$ shows how much the global layout of a map depends on its seed.} On the MNIST parameter paths each setting was run three times; on the t-SNE paths seed 0 starts from PCA and seeds 1 and 2 at random, so the spread includes the change of initialization. The refits of Supplementary Note~\ref{app:refit} all start at random and move the global difference by 0.010. The standard deviation of $G$ over the runs exceeds that of $L$ in 23 of the 24 settings, by a median factor of 34 (Supplementary Fig.~\ref{fig:robust}b). At perplexity 30 it is 0.038 for $G$ and 0.0005 for $L$, a factor of 84. Over all settings the standard deviation of $L$ is at most 0.0017, and that of $G$ reaches 0.128 with two principal components before t-SNE. The nearest neighbors in a stochastic map are reproducible across runs, and the global layout depends on the random start. We therefore report the run-to-run standard deviation of $G$ wherever runs are repeated (Supplementary Note~\ref{sec:findings}).

\runhead{Agreement rises with the scale.} With the distance rank correlation it rises from $-0.18$ at $k=10$ to 0.57 at the middle band and 0.84 to 0.94 at the upper end. With random-triplet accuracy it rises from $-0.10$ to 0.70 and 0.93 to 0.98.

\subsection{Four applications: pathology, single cells, supply chains, and AI agent memory}
\runhead{Named axes.} NPCA names its axes on the three datasets that have class labels or named variables (Figure~\ref{fig:appsbig}d,i,n). On Pathology-100K the axes are debris and muscle, which hold 36\% and 49\% of the 5\% of tiles scoring highest on them. On SEA-AD they are oligodendrocytes, 54\%, and microglia with perivascular macrophages, 58\%. On Instacart they are two baskets: organic bananas, organic baby spinach and organic strawberries on one, and bananas, large lemons and organic avocados on the other. Named axes keep $G$ within 0.002 of PCA's on SEA-AD and Instacart, and within 0.136 on Pathology-100K.

\runhead{Smaller datasets.} The three smaller datasets show the same pattern at a few thousand points (Supplementary Fig.~\ref{fig:apps}). They are Pathology-7K, Online Retail and 8-K described in Methods; the filings serve as the long-term memory of a finance agent. PCA has the highest $G$ on Pathology-7K and 8-K, 0.549 and 0.440. On Online Retail t-SNE and SR-tSNE2 are the most global maps, with $G=0.191$ against 0.103 for PCA, as on Instacart. NPCA's axes are two tissues on Pathology-7K, background and normal mucosa, and two product groups on Online Retail, with the $G$ of PCA. On 8-K they are two filing types, earnings releases (item 2.02, 100\%) and officer changes (item 5.02, 96\%).

\runhead{What $L$ means for a pathologist.} A difference in $L$ has a direct reading for a user. On the SR-tSNE2 map a tumor tile keeps 6.1 of its 10 nearest neighbors, and on UMAP 4.1 (Supplementary Fig.~\ref{fig:path}). The SR-tSNE2 map separates the nine tissues and keeps the morphology continuous across them (Supplementary Fig.~\ref{fig:path}a,b). The NPCA axes are tissues, and the tiles that score highest on them are the tissue they name (Supplementary Fig.~\ref{fig:path}c). These counts are averages over all 1,233 tumor tiles, with neighbors defined in the Phikon space; for SR-tSNE2 the 95\% bootstrap interval is 6.0 to 6.2. The t-SNE map keeps 6.0, UMAP 4.1, PaCMAP 3.7 and PCA 0.4. SR-tSNE2 returns at least half of the true neighbors for 80\% of the tumor tiles, and UMAP for 41\%. Supplementary Fig.~\ref{fig:path}d shows one tumor tile drawn at random. For a pathologist who searches a map for similar tissue, the map with the higher $L$ returns more of the true neighbors.

\subsection{What $L$ and $G$ show about common methods}\label{sec:findings}

SR-tSNE2 is the most local map on every large dataset, the most global map changes with the data, and across all 24 MNIST parameter settings PCA keeps the highest $G$ and t-SNE the highest $L$.

\runhead{Which method is global depends on the data at every size tested.} The four applications showed that the most global map changes from one dataset to the next. Supplementary Table~\ref{tab:largeLG} and Supplementary Fig.~\ref{fig:LGplane} give the scores on all twelve datasets. SR-tSNE2 has the highest $L$ on all twelve large datasets, by 0.0002 to 0.056 over t-SNE, and t-SNE the highest among the published methods. A one-sided test on all points finds the gain of SR-tSNE2 significant on nine datasets, by 0.010 to 0.056 with $p\le3.7\times10^{-5}$, and within 0.001 on CIFAR-10, Speech Commands and AG News (Supplementary Note~\ref{app:localfull}). The highest $G$ belongs to PCA on six of the twelve: MNIST, Fashion-MNIST, Speech Commands, AG News, Pathology-100K and MS MARCO. PaCMAP has the highest $G$ on Brain-Cortex, Covertype, SEA-AD and Instacart, 0.413 against 0.328 for PCA on Brain-Cortex and 0.417 against 0.286 on SEA-AD. NPCA leads by 0.002 on Renal-Cortex and UMAP by 0.001 on CIFAR-10. PCA has the highest $G$ on half of the large datasets.

\runhead{On large data, PaCMAP leads at the upper end and t-SNE at the lower end.} PaCMAP was designed to preserve both local and global structure \citep{wang2021}. It has a higher $G$ than t-SNE on seven of the twelve large datasets, by up to 0.174 on SEA-AD, and equals it on AG News. UMAP is above t-SNE on four. At the lower end t-SNE leads both: its $L$ is 1.6 to 11 times that of UMAP and 2.1 to 26 times that of PaCMAP on every large dataset. Among the published methods we tested, the leaders at the two ends are different maps, and the reversal test declares t-SNE local and PaCMAP global on seven datasets (Supplementary Note~\ref{sec:pairs}).

\runhead{On the Swiss roll and the S-curve the neighbor embeddings are global.} PCA keeps the large Euclidean distances by construction, and the large datasets use a Euclidean or cosine reference. The three synthetic manifolds have a known geometry and a geodesic reference (Supplementary Fig.~\ref{fig:synth} and Supplementary Table~\ref{tab:synth}). PCA has $G$ of 0.24 and 0.29 on the Swiss roll and the S-curve, where t-SNE reaches 0.84 and 0.85 and Isomap, whose objective is the geodesic distance, 0.98 and 0.90. SR-tSNE2 has the highest $L$ on both, 0.93 and 0.82. NPCA matches PCA on the Swiss roll, and PCA leads it on the S-curve, 0.29 against 0.13. The PCA map of the Swiss roll overlays the turns of the roll that the neighbor embeddings unroll (Supplementary Fig.~\ref{fig:synth}g--k). On the tree PCA's $G$ of 0.53 is close to t-SNE's 0.55, and Isomap has the highest, 0.60. The claim that PCA is global holds when the large-scale structure of the data is close to linear, and on manifolds that roll up the neighbor embeddings take its place.

\runhead{The scores measure the design choices of our two methods.} The same scale applies to new methods. Supplementary Table~\ref{tab:own} compares each of our two methods with the method it modifies on the twelve large datasets. SR-tSNE2 returns the 10 nearest neighbors at their exact rank 18\% to 76\% more often than the t-SNE map it refines, and by more than 35\% on eight (Supplementary Fig.~\ref{fig:own}d). Its $L$ rises by up to 0.056 on MS MARCO, and its $G$ stays within 0.0005 of that of t-SNE (Supplementary Fig.~\ref{fig:own}e). On large data the rank refinement improves local order without changing $G$, and the reversal test declares SR-tSNE2 local on the same datasets as t-SNE. NPCA keeps every loading nonnegative, so each axis is an additive combination of the input variables. Its $G$ is within 0.005 of PCA's on Renal-Cortex, SEA-AD and Instacart, and PCA leads it by 0.006 to 0.136 on the other nine (Supplementary Fig.~\ref{fig:own}c). NPCA thus gains interpretable axes, with $G$ close to PCA's on three datasets. The reversal test declares t-SNE local and NPCA global on four datasets (Supplementary Note~\ref{sec:pairs}). Supplementary Fig.~\ref{fig:own}a,b shows how the settings were chosen on the 60,000 MNIST training digits. With learning rate 50 and 2,000 iterations, a rank weight $\lambda=20$ raises O-DLP@10 by 29\% and $L$ by 0.007, and $\lambda=50$ lowers $L$ by 0.030.

\runhead{Parameter paths have measurable exchange rates.} Across all 24 settings of four parameter paths, PCA keeps the highest $G$ and t-SNE the highest $L$. Besides the choice of method, each method has parameters that are said to move a map toward global structure. Supplementary Fig.~\ref{fig:tradeoff} traces four parameter paths in the $(L,G)$ plane on the 60,000 MNIST training digits, and Supplementary Table~\ref{tab:sweep} lists them with their energy. $L$ is computed exactly over all points and $G$ from ten subsamples of 2,000. Each point is the mean of three runs, one with PCA initialization and two random.

\runhead{Perplexity.} Raising t-SNE's perplexity from 30 to 500 lowers $L$ by 0.190 and raises $G$ by 0.051, an exchange rate of 0.27. It costs 6.2 times the energy, because t-SNE computes its sparse affinities over three times the perplexity in neighbors (Supplementary Fig.~\ref{fig:tradeoff}b). From 30 to 200, the rate is 0.33 at 3.4 times the energy; on the 70,000 digits of Figure~\ref{fig:concept}, from a single run, it is 0.21. At perplexity 500 t-SNE's $G$ is 0.269 and PCA's 0.318: at 60,000 points PCA stays the global map across the perplexities we tested. Whether t-SNE is local thus depends on its perplexity, and on large data each gain in $G$ comes at an estimated energy.

\runhead{Neighbors, exaggeration and preprocessing.} For every UMAP setting, some perplexity of t-SNE has both a higher $L$ and a higher $G$. From 5 to 500 neighbors, UMAP's $L$ moves from 0.151 to 0.063 and its $G$ stays within 0.018. Exaggeration moves t-SNE along the attraction--repulsion spectrum toward UMAP-like maps \citep{bohm2022}. From exaggeration 1 to 32, $L$ moves from 0.370 to 0.045 and $G$ stays within 0.011. Stronger attraction tightens the clusters and leaves their layout almost unchanged. PCA preprocessing interpolates between the two maps. Five components give the highest $G$ of all 24 settings, 0.311, within 0.007 of PCA, at $L=0.084$. Two components raise the run-to-run spread of $G$ to 0.128, larger than any difference between methods.

\subsection{The reversal test at scale}

\subsubsection{A worked example: the memory of a retrieval agent}
\runhead{Two kinds of question.} An agent keeps one million MS MARCO passages as BGE-base vectors and retrieves in that 768-dimensional space. A two-dimensional map is the view in which a person or the agent inspects and organizes the memory, and it serves two kinds of question. One asks which passages lie near a query and its retrieved set, a question about a few points. The other asks which topics lie far apart and which to merge or delete, a question about large parts of the memory.

\runhead{Which local map.} SR-tSNE2 is the map for the passages nearest a query. At the exact 10 nearest neighbors of all one million passages it leads t-SNE by 0.056 ($p=1.2\times10^{-21}$, Supplementary Note~\ref{app:localfull}), and on the subsample of 5,000 passages t-SNE leads it by 0.0008.

\subsubsection{The test on independent samples}\label{sec:sim}
The maps of each large dataset were kept fixed, and the test was rerun on 50 independent uniform subsamples of 5,000 points (Figure~\ref{fig:sim}, Supplementary Table~\ref{tab:sim}). For t-SNE against PCA the declaration is the same in all 50 subsamples of every dataset, and the crossover moves by a standard deviation of 0.002 to 0.008. Its jackknife interval covers the mean crossover in every subsample on the eleven datasets with a crossover inside the data. Over six pairs of maps on twelve datasets, the local test rejects a true null at level 0.05 at no more than 0.020 and the global test at 0. A test that treats the points as independent rejects at up to 0.20 and 0.48. In this experiment the null for the global test is centered at the full-data difference in $G$, which is read on a band of relative scale, and for the local test at the mean over the subsamples, because $L$ is read at $k=10$ and so depends on $n$.

\subsubsection{Neighbor embeddings against PaCMAP and NPCA}\label{sec:pairs}

Against PaCMAP, t-SNE and SR-tSNE2 are declared local on seven of the twelve large datasets, and on the other five they lead PaCMAP at both ends. The test compares any two maps, and PaCMAP has the highest $G$ on four large datasets (Supplementary Table~\ref{tab:largeLG}). Supplementary Table~\ref{tab:pairs} sets UMAP, t-SNE and SR-tSNE2 against PaCMAP with the procedure used against PCA. Each of the three is more local than PaCMAP on every dataset, with $p_L\le0.021$. The test declares t-SNE local and PaCMAP global on Pathology-100K, SEA-AD, Instacart, CIFAR-10, Speech Commands, Brain-Cortex and Covertype, the seven datasets on which PaCMAP has the higher $G$ in Supplementary Table~\ref{tab:largeLG}. The test and the scores of Supplementary Table~\ref{tab:largeLG}, computed on separate subsamples, thus agree on all twelve datasets. SR-tSNE2 is declared local on the same seven, and UMAP on six, MS MARCO among them. For SR-tSNE2 the crossover lies at 0.4\% to 8.4\% of the data, below its crossover with PCA on every dataset, so PaCMAP becomes the better map at smaller question sizes than PCA does. On MS MARCO, MNIST, Fashion-MNIST, Renal-Cortex and AG News, the $G$ of t-SNE and SR-tSNE2 in the test is 0.002 to 0.071 above that of PaCMAP, and no reversal is declared.

NPCA, whose axes are additive combinations of the input variables, takes the place of PCA as the global map on three datasets, and at the level on a fourth. The test declares t-SNE local and NPCA global on Fashion-MNIST, Speech Commands and Covertype, and at the level on AG News ($p=0.04999$), and SR-tSNE2 on the first three, each more local with $p_L\le3.8\times10^{-9}$. The crossover lies at 11\% to 32\% of the data, 1.1 to 1.7 times that with PCA. On the other eight datasets the global gap stays within its sampling error or favors t-SNE. Set directly against PCA on these subsamples of 5,000 points, NPCA matches it at the upper end on Instacart, Renal-Cortex and Brain-Cortex, with gaps within their sampling error, and comes within 0.002 on SEA-AD. On the other eight PCA leads by 0.012 to 0.129, and NPCA offers named axes at that cost. On a subsample of 10,000 points the same eight gaps are significant. Over these 60 one-sided global tests the rejection rate under the null averages 0.052 at level 0.05 and ranges from 0.010 to 0.128. The test thus holds its level with PaCMAP or NPCA as the global map, as it does with PCA.

\subsubsection{A stable PCA map}\label{sec:pcastab}
The stability of the PCA map explains this dependence. PCA gives a stable map when its plane is well determined. The test compares two given maps, so a map that changes when it is refitted changes the declaration. PCA changes in this way when the second and third eigenvalues of the covariance are close. A wide gap between them fixes the plane of the first two principal axes. A perturbation of the data turns it by an angle bounded by the size of the perturbation divided by the eigenvalue gap \citep{davis1970}. On the 70,000 MNIST points the two eigenvalues differ by 8\% (ratio 1.08). The sampling perturbation shrinks as $1/\sqrt n$, so at 70,000 points an 8\% gap fixes the plane, and at 5,000 points a larger gap is needed.

We drew 40 random subsets of 5,000 MNIST points and fitted PCA on each (Supplementary Fig.~\ref{fig:scale}d,e). $L$ of the PCA map stays at 0.040 (standard deviation 0.005), and $G$ splits in two. In 30 of them the plane stays close to the full-data plane and $G$ averages 0.303. In the other 10 the second axis turns by an angle $\theta_2$ of more than 60 degrees ($\cos\theta_2<0.5$), and $G$ averages 0.162. PCA fitted on the subset is more global than t-SNE in 31 of the 40 subsets. The full-data plane, applied to the same subsets, has $G=0.303$ (standard deviation 0.004) and is more global than t-SNE in all 40.

\runhead{Checking the plane.} A turning plane is best detected by refitting it: four MNIST subsets of 5,000 points with eigenvalue ratios from 1.11 to 1.32 all turn. The function \texttt{pca\_stability} of \texttt{lgfid} (command \texttt{lgfid pca DATA}) refits the plane on 10 random halves of the data and reports the smallest cosine between a refit and the full-data plane. The smallest cosine is 0.001 to 0.067 on those four subsets and 0.996 on all 70,000 points. A cosine of 0.9 or more marks a stable plane; below it, the user compares maps against the PCA plane of the largest available sample, which the command saves.

\subsubsection{What the test tells a user}\label{sec:lguse}

Taken together, these results answer four questions a user has about the four applications (Supplementary Table~\ref{tab:lguse}).

\runhead{One map or two.} A declared reversal means that each map answers one kind of question better, and the user needs two maps. A pair in which one map is more faithful at both ends needs one map. A pair without a declared reversal gives no evidence that two maps are needed, and the estimates of $L$ and $G$ show which map leads. On Online Retail t-SNE is more local than PCA by 0.239 and more global by 0.088, and its $G$ is the highest of all maps, shared with SR-tSNE2. One t-SNE map therefore serves both the search for customers who buy alike and the layout of customer segments.

\runhead{When to change maps.} The crossover gives the size of question at which the user changes maps. A question about fewer than $s^\times n$ points, such as the nearest neighbors of one point or a small subtype, goes to the local map. A question about more points, such as the layout of classes, goes to the global map. On Pathology-100K the crossover lies at about 10,000 tiles, and on Pathology-7K at 776 tiles, the same tenth of the data.

\runhead{How to compare clusters.} The distance between two clusters that together hold more points than the crossover size is read from the global map, and that between two small clusters from the local map. The p-value of the declaration gives the evidence for this choice on the data at hand.

\runhead{Whether the usual claim holds.} The test checks the claim that t-SNE is local and PCA global on the user's data. It holds firmly on six of the twelve large datasets, with 10, 20 and 50 blocks, and narrowly on MNIST, CIFAR-10 and Renal-Cortex. On the Swiss roll, the S-curve, Instacart and Online Retail the neighbor embeddings are the more global maps, and on SEA-AD the global gap stays within its sampling error. The command \texttt{lgfid test DATA A B} returns the declaration, its p-value, and the crossover with its interval for any two maps of the user's data.

\runhead{Perplexity.} Raising the perplexity from 30 to 200 lowers PCA's global advantage over t-SNE on nine of the twelve datasets, and the reversal stays declared on six. A larger neighborhood moves t-SNE toward the upper end, as the perplexity path of Supplementary Fig.~\ref{fig:tradeoff} shows.

\subsection{Memories of artificial agents}

The scale also applies when the memory is text. In a pilot on four LoCoMo conversations \citep{maharana2024locomo}, Qwen2.5-14B read a compressed context of about a quarter of the conversation's tokens and restated each turn from the turn before it. We embedded the restated turns and compared them with the originals. Five compressions, among them the dataset's session summaries, lowered $L$ from 0.57 to 0.05--0.08 and $G$ from 0.69 to 0.09--0.15, close to the values of an empty context (0.04 and 0.07). Compressing the text of a conversation loses both its local and its global structure.

PCA and NPCA place a new item by projection, and openTSNE places one on a t-SNE map \citep{policar2024}. SR-tSNE2 places a new item on the t-SNE map it refines and refines it with the map at the next refit.

\FloatBarrier
\section{Supplementary figures}\label{app:figs}
\setcounter{figure}{0}\renewcommand{\thefigure}{\arabic{figure}}

\begin{figure}[!htbp]
\centering
\includegraphics[width=\textwidth]{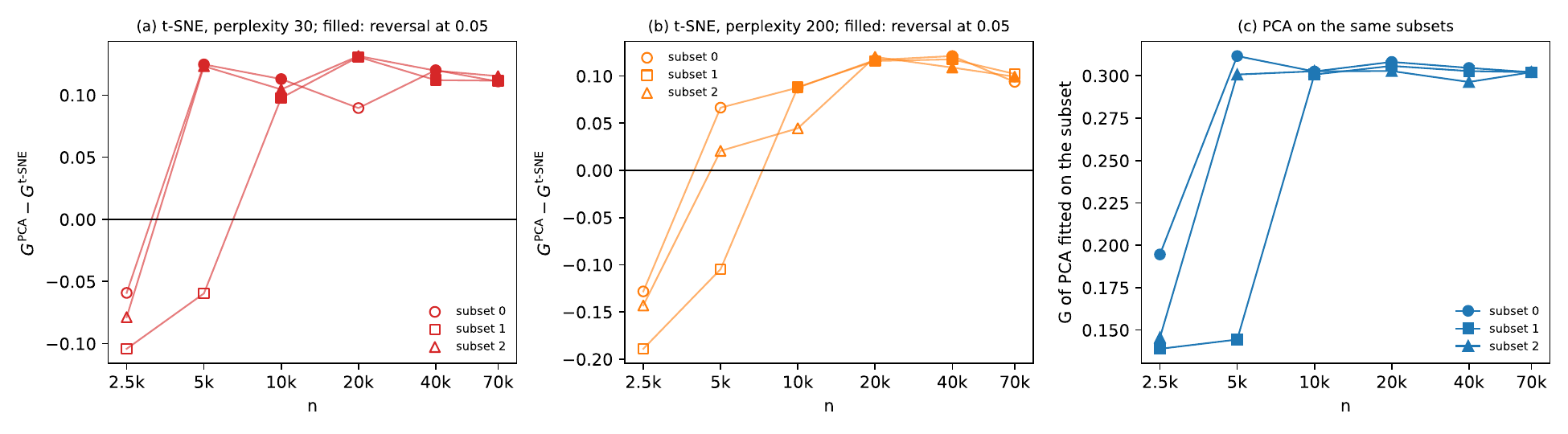}\\[4pt]
\includegraphics[width=0.75\textwidth]{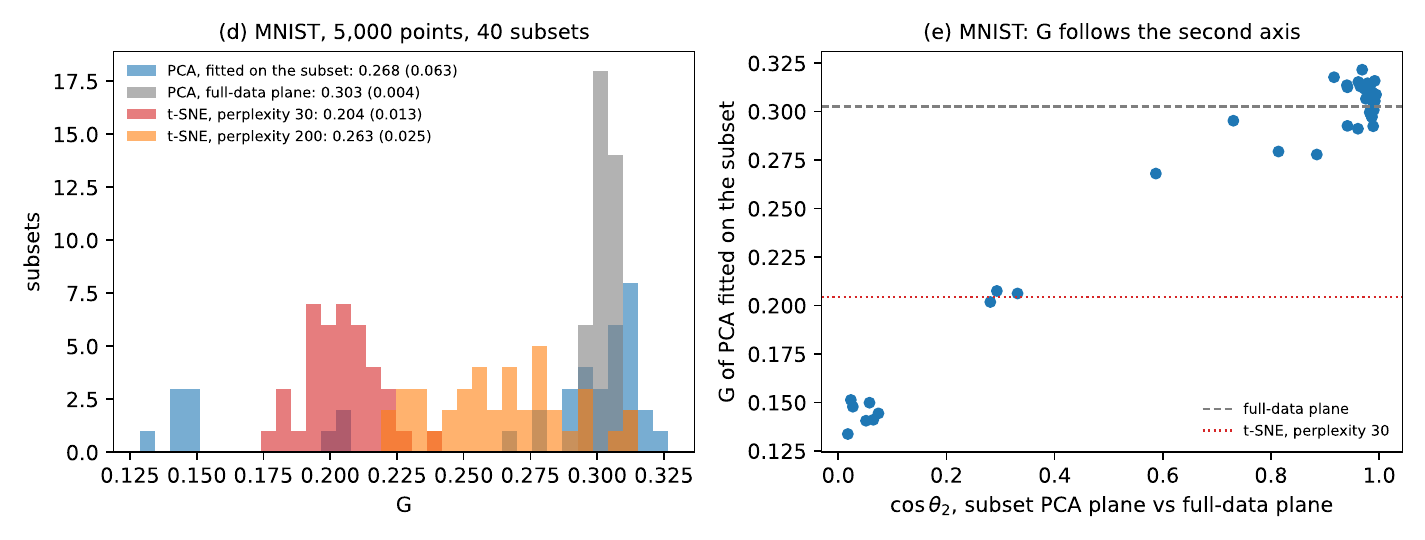}
\caption{From 10,000 points PCA's two axes are stable and the reversal is declared in at least two of three nested series, and in all three at 40,000 and 70,000. (a), (b) $G^{\rm PCA}-G^{\rm t\text{-}SNE}$ on nested MNIST subsets at perplexity 30 and 200; filled markers, reversal declared. (c) $G$ of PCA fitted on each subset. (d), (e) Forty subsets of 5,000 points: the full-data plane keeps $G$ near 0.30 in all of them.}\label{fig:scale}
\end{figure}

\begin{figure}[!htbp]
\centering
\includegraphics[width=\textwidth]{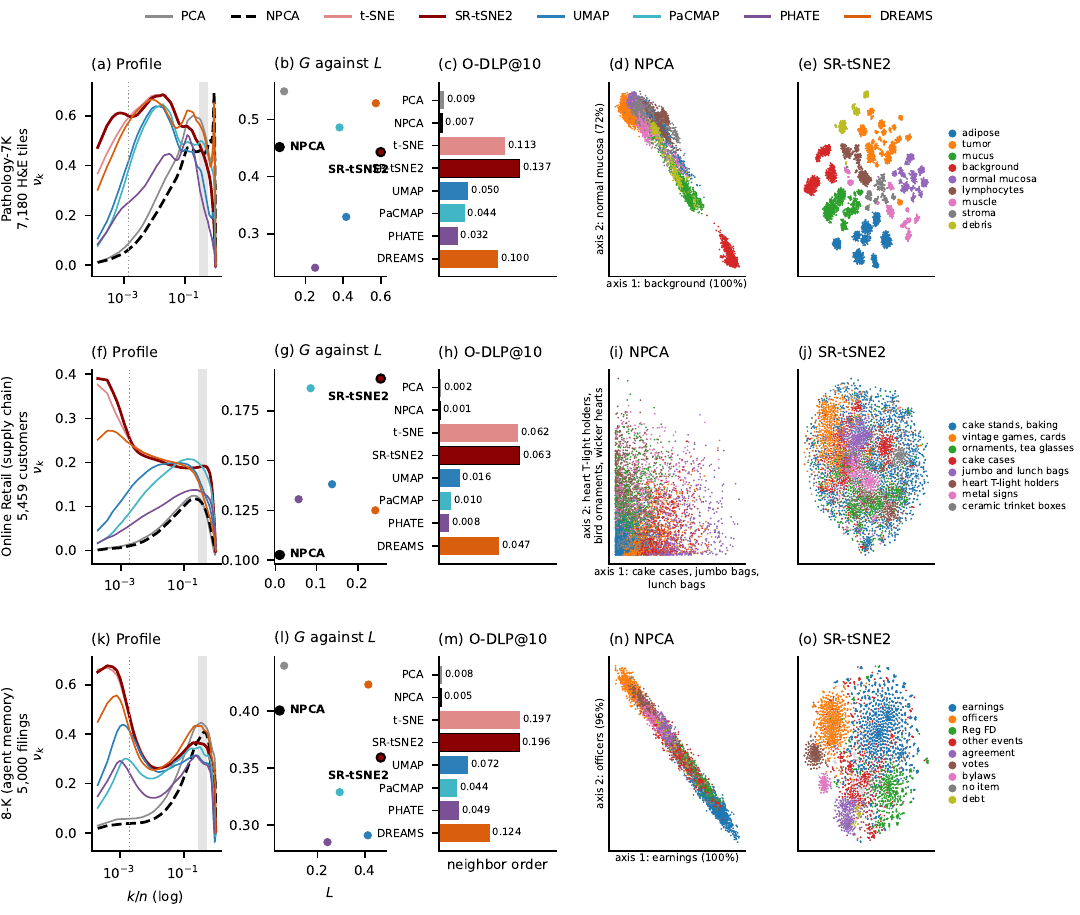}
\caption{On three smaller datasets PCA is the most global map of Pathology-7K and 8-K, and t-SNE and SR-tSNE2 the most global of Online Retail. Columns as in Figure~\ref{fig:appsbig}, with PHATE and DREAMS \citep{kury2026} added and t-SNE$_{200}$ omitted; the SR-tSNE2 map is colored by tissue, purchase cluster or main 8-K item.}\label{fig:apps}
\end{figure}

\begin{figure}[!htbp]
\centering
\includegraphics[width=\textwidth]{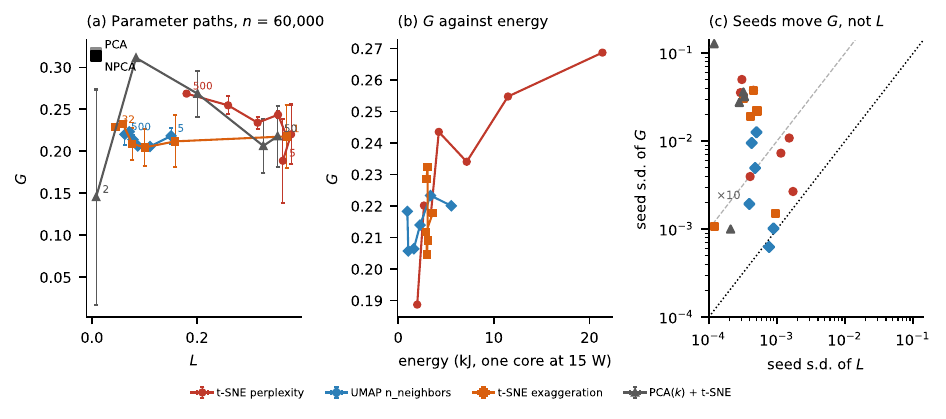}
\caption{On 60,000 MNIST digits PCA keeps the highest $G$ and t-SNE the highest $L$ across all parameter settings. (a) Four parameter paths in the $(L,G)$ plane, mean $\pm$ one standard deviation over three runs; squares, PCA and NPCA. (b) Energy along each path. (c) Run-to-run standard deviation of $G$ against that of $L$; dotted line $y=x$, dashed line $y=10x$.}\label{fig:tradeoff}
\end{figure}

\begin{figure}[!htbp]
\centering
\includegraphics[width=\textwidth]{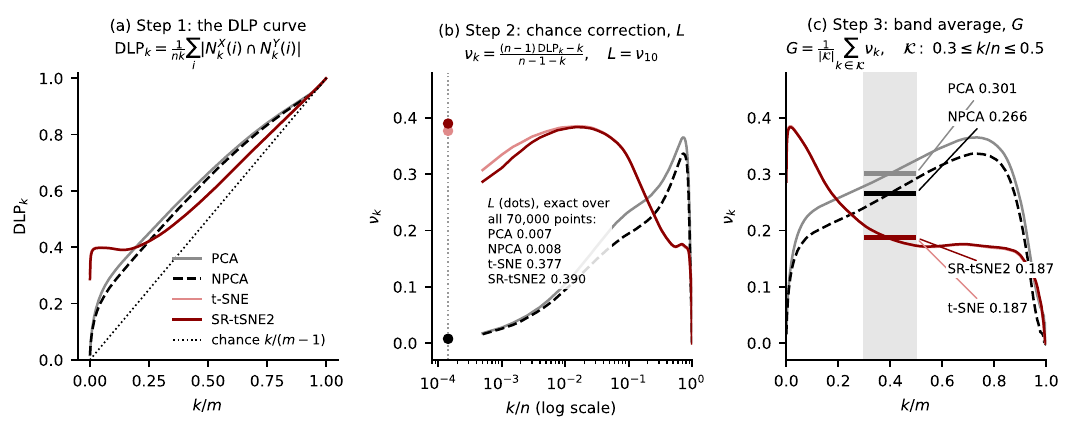}
\caption{The chance correction reveals the reversal on MNIST: SR-tSNE2 is the most local of four maps and PCA the most global. (a) $\DLP_k$ nears its chance level at large $k$. (b) The profile $\nu_k$; dots mark $L$. (c) Over the shaded band, which gives $G$, the order reverses.}\label{fig:LGcalc}
\end{figure}

\begin{figure}[!htbp]
\centering
\includegraphics[width=\textwidth]{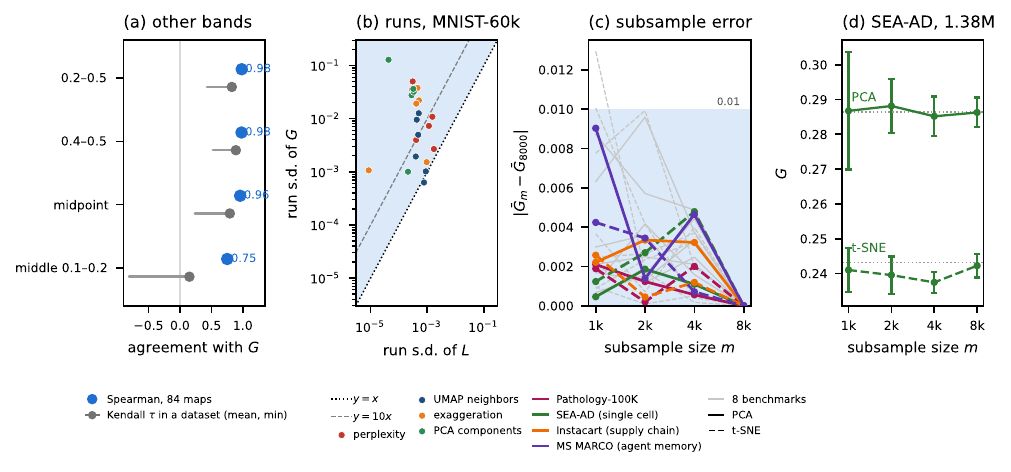}
\caption{$G$ is stable across bands and subsample sizes, and it varies across runs more than $L$. (a) Correlation of three other bands with $G$; gray, agreement of the method rankings within a dataset. (b) Run-to-run standard deviation of $G$ against that of $L$; dotted line $y=x$. (c) Change in $G$ with the subsample size $m$, PCA solid and t-SNE dashed; colored, the four applications. (d) $G$ on SEA-AD from $m=1{,}000$ to 8,000.}\label{fig:robust}
\end{figure}

\begin{figure}[!htbp]
\centering
\includegraphics[width=\textwidth]{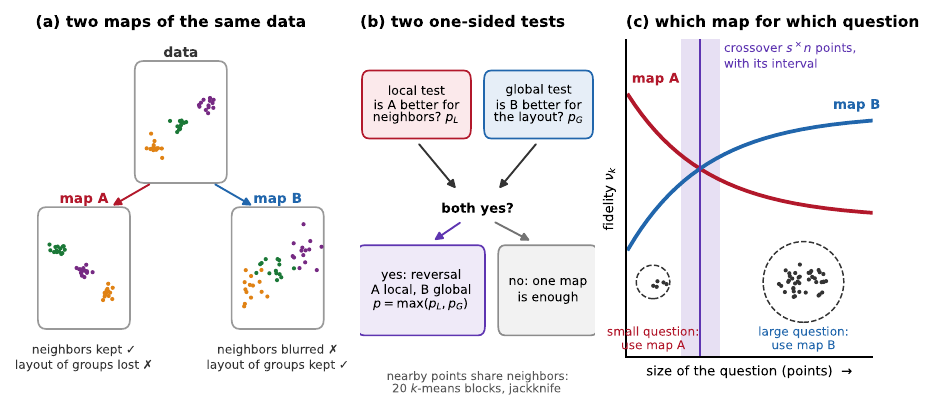}
\caption{The LG reversal test tells a user which of two maps answers which question (schematic). (a) Map A keeps the neighbors; map B keeps the layout of the groups. (b) A reversal is declared when both one-sided tests reject. (c) Questions about fewer points than the crossover go to map A, larger ones to map B.}\label{fig:cartoon}
\end{figure}

\begin{figure}[!htbp]
\centering
\includegraphics[width=\textwidth]{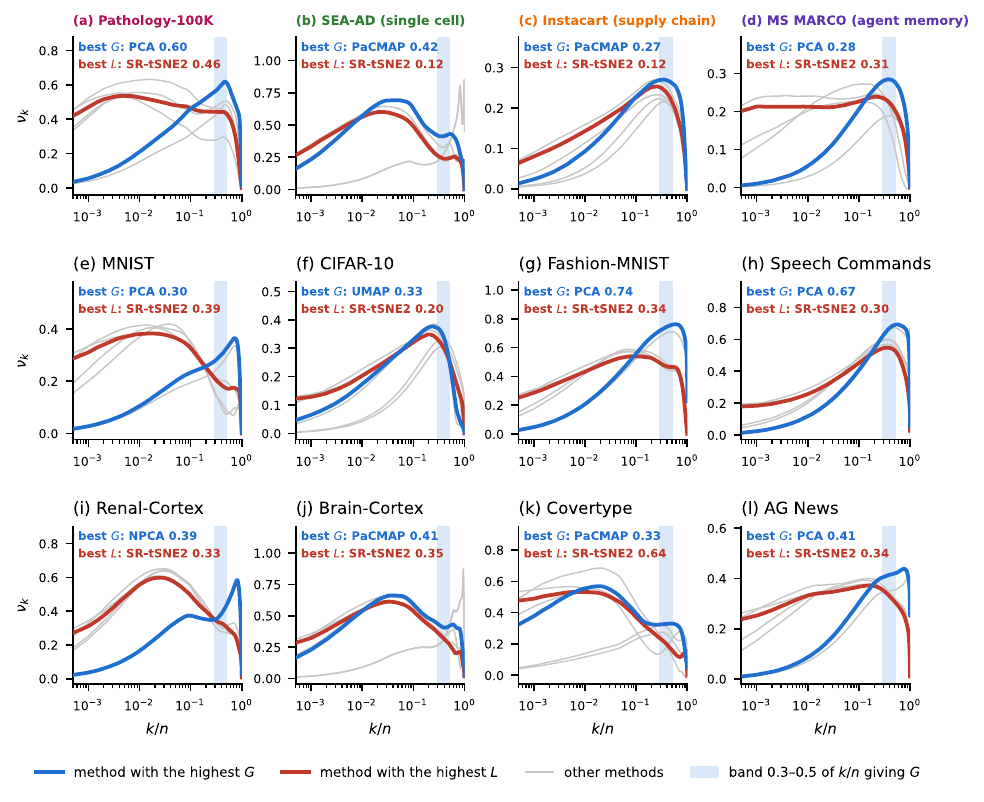}
\caption{Which method is most faithful depends on the scale and the data. Profiles of seven methods on the twelve large datasets, the four applications in the top row; blue, highest $G$; red, highest $L$; shaded band, the range that gives $G$.}\label{fig:profiles}
\end{figure}

\begin{figure}[!htbp]
\centering
\includegraphics[width=0.96\textwidth]{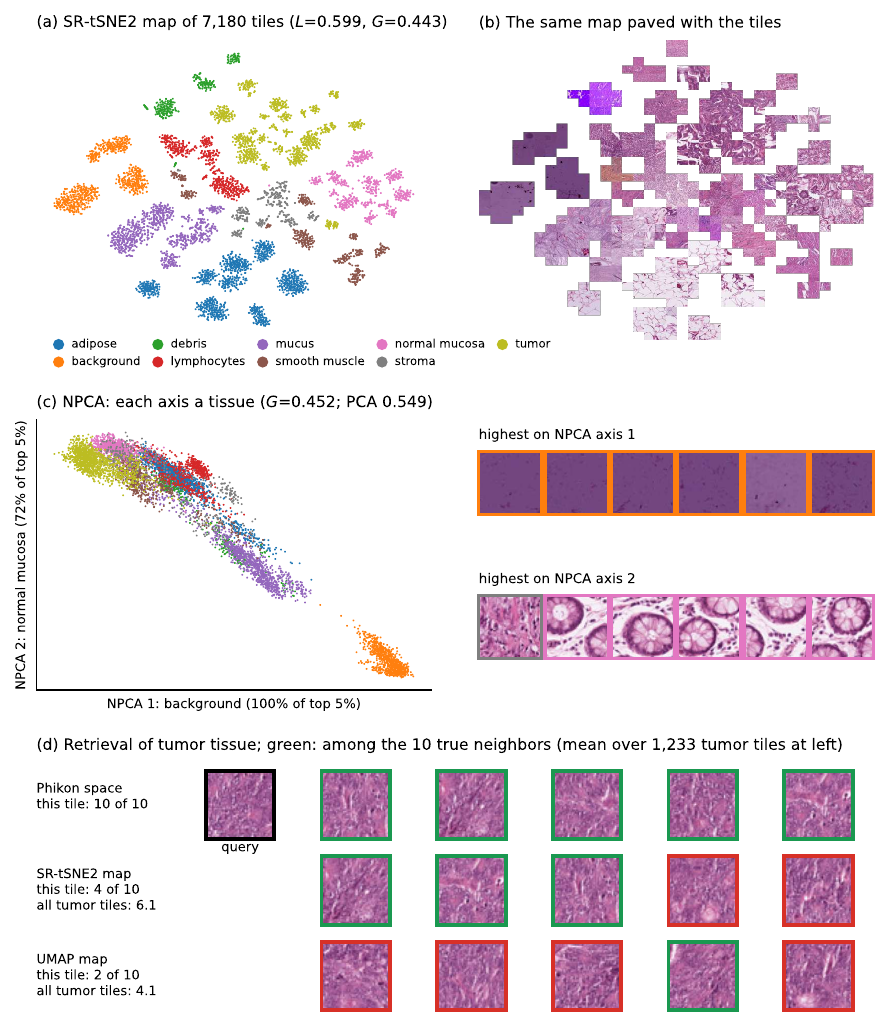}
\caption{On the SR-tSNE2 map a tumor tile keeps 6.1 of its 10 true neighbors on average, and on UMAP 4.1. (a) The SR-tSNE2 map of Pathology-7K by tissue. (b) The same map, showing the tile nearest each grid cell. (c) The NPCA map with the six top tiles on each axis. (d) A tumor tile and its five nearest neighbors on each map; green frames mark true neighbors in the Phikon space.}\label{fig:path}
\end{figure}

\begin{figure}[!htbp]
\centering
\includegraphics[width=\textwidth]{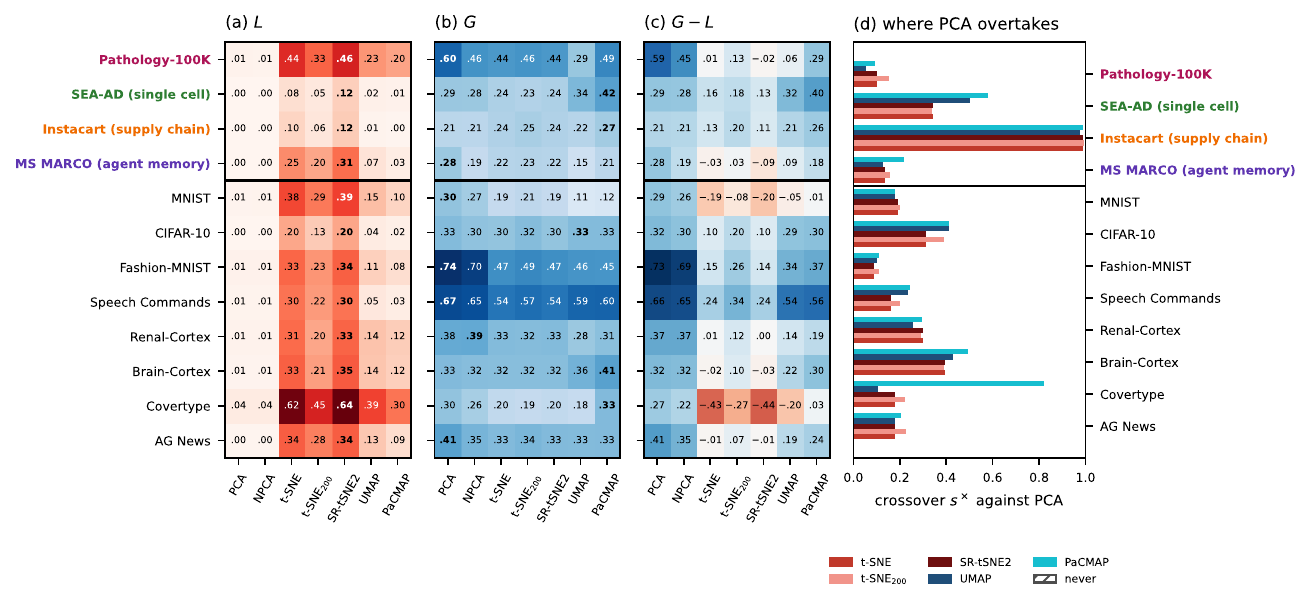}
\caption{SR-tSNE2 is the most local map on every large dataset, and the most global map changes with the data. Applications above the line. (a) $L$. (b) $G$; bold, the best. (c) $G-L$, compared between maps of one dataset; blue, positive. (d) Crossover scale at which PCA reaches each neighbor embedding.}\label{fig:LGplane}
\end{figure}

\begin{figure}[!htbp]
\centering
\includegraphics[width=\textwidth]{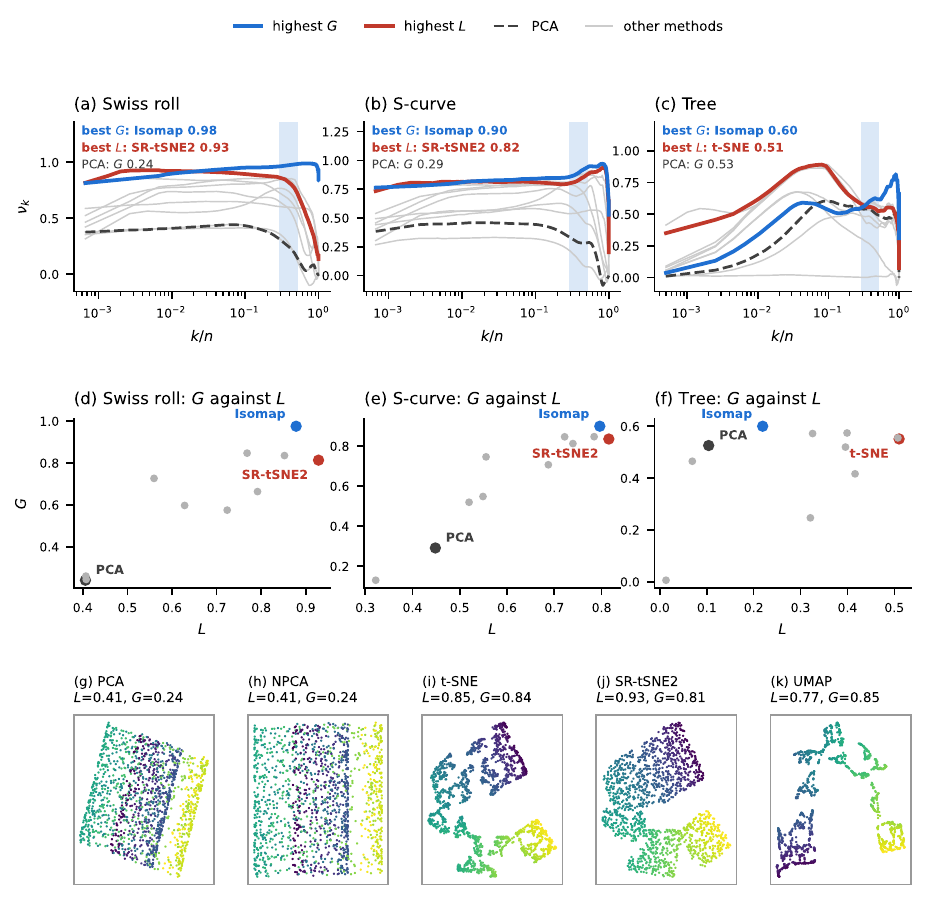}
\caption{On the Swiss roll and the S-curve the neighbor embeddings unroll the manifold and are far more global than PCA. The reference is the geodesic distance. (a)--(c) Profiles of eleven methods; blue, highest $G$; red, highest $L$; dashed gray, PCA. (d)--(f) $G$ against $L$. (g)--(k) Swiss-roll maps colored by position along the roll: PCA and NPCA overlay its turns.}\label{fig:synth}
\end{figure}

\begin{figure}[!htbp]
\centering
\includegraphics[width=\textwidth]{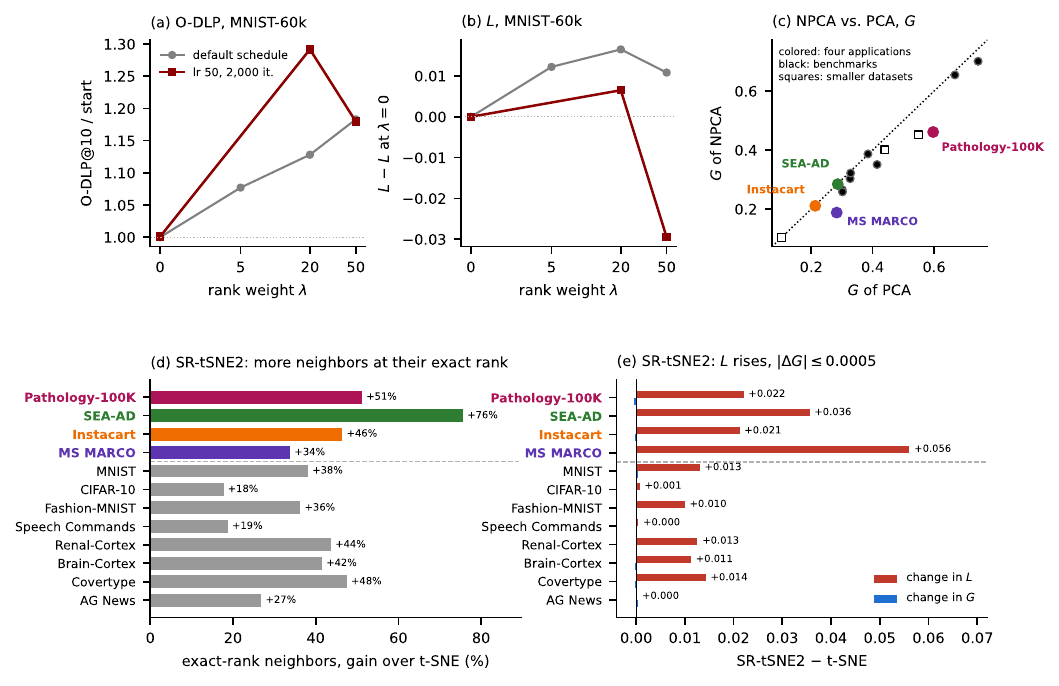}
\caption{SR-tSNE2 improves local order with $G$ within 0.0005, and NPCA gains named axes with $G$ close to PCA's on three datasets. (a), (b) Tuning the rank weight $\lambda$ on 60,000 MNIST digits. (c) $G$ of NPCA against $G$ of PCA; colored, the applications; squares, the smaller datasets. (d), (e) SR-tSNE2 against the t-SNE map it refines: (d) gain in O-DLP@10; (e) change in $L$ and $G$.}\label{fig:own}
\end{figure}

\FloatBarrier
\sitablesbegin
\section{Supplementary tables}\label{app:tabs}
\setcounter{table}{0}\renewcommand{\thetable}{\arabic{table}}

\begin{table}[!htbp]
\centering\footnotesize
\setlength{\tabcolsep}{3pt}
\caption{Our DLP curve \citep{han2022} is the curve from which every score here is read. Earlier work supplies the global scores we use to validate $G$ and the other curves listed here. This paper adds five features. One statistic: the local and global scores are one statistic at two scales. Relative band: the global score uses a fixed range of $k/n$. Proven meaning: a proof of what the upper end of the score measures. Rate: a crossover or an exchange rate. Test: a test of whether one map is local and another global.}\label{tab:novelty}
\begin{tabular}{@{}lllccccc@{}}
\toprule
Work & Local score & Global score & \shortstack{One\\statistic} & \shortstack{Relative\\band} & \shortstack{Proven\\meaning} & Rate & Test \\
\midrule
Our DLP curve \citep{han2022} & \multicolumn{2}{l}{$\DLP_k$ over all $k$, the curve every score here is read from} & & & & & \\
\citet{lee2009} & $Q_{\text{local}}$, mean $Q_{NX}$, $k\le k_{\max}$ & $Q_{\text{global}}$, mean $Q_{NX}$, $k>k_{\max}$ & \ding{51} & & & & \\
\citet{lee2015} & \multicolumn{2}{l}{$R_{NX}(k)$ over all $k$, area on a $\log k$ axis} & \ding{51} & & & & \\
DREAMS \citep{kury2026} & $k$NN recall, $k=10$ & Spearman $\rho$ of distances & & & & & \\
This paper & $L=\nu_{10}$ & $G$, mean $\nu_k$, $0.3\le k/n\le0.5$ & \ding{51} & \ding{51} & \ding{51} & \ding{51} & \ding{51} \\
\bottomrule
\end{tabular}
\end{table}

\begin{table}[!htbp]
\centering\small
\caption{$G$ agrees with the established global scores at 0.84 and 0.93, and the area under $R_{NX}$ at 0.22 and 0.30. Spearman rank correlation between profile-based scores and established global scores over the 84 maps of the large datasets (twelve datasets, seven methods).}\label{tab:validation}
\begin{tabular}{@{}lcc@{}}
\toprule
Profile score & Distance $\rho$ & Random-triplet accuracy \\
\midrule
\csname @@input\endcsname tab_validation_rows 
\bottomrule
\end{tabular}
\end{table}

\begin{table}[!htbp]
\centering\small
\caption{Notation.}\label{tab:notation}
\begin{tabular}{@{}ll@{}}
\toprule
Symbol & Meaning \\
\midrule
$\x_i$, $\y_i$, $n$ & data point, its position in the map, number of points \\
$d_X$, $d_Y$ & reference dissimilarity of the data; Euclidean distance in the map \\
$r^X_i(j)$, $r^Y_i(j)$ & rank of point $j$ among the other points by distance from point $i$ \\
$N^X_k(i)$, $N^Y_k(i)$ & the $k$ nearest neighbors of point $i$ in the data and in the map \\
$\DLP_k$, $\nu_k$ & degree of locality preservation at $k$, and its chance-corrected value \\
$h(i)$, $h$ & halves agreement of point $i$, and its mean over points \\
$s=k/n$ & relative neighborhood size (relative scale) \\
$k^\star=(n-1)/2$ & neighborhood size of the near half of the data \\
$L$, $G$ & local fidelity $\nu_{10}$; global fidelity, mean of $\nu_k$ over $0.3\le k/n\le0.5$ \\
$s^\times(A,B)$ & crossover scale of method $A$ against method $B$ \\
$\eta$ & exchange rate of a parameter path \\
$L^A$, $\nu^A_k$, $L(i)$ & a score of method $A$; a score of one point $i$ \\
$O_k(i)$, $\nu_k(i)$ & neighbors of point $i$ kept at $k$; its chance-corrected profile \\
$E(i)$, $E$, $M$ & entity fidelity of point $i$ and of the map; the profile at the median entity size \\
$n_{c(i)}$, $k_i$ & size of the entity of point $i$; $k_i=n_{c(i)}-1$ \\
$d^L(i)$, $d^G(i)$ & per-point local and global differences between two maps \\
$C$, $p_L$, $p_G$, $\alpha$ & number of jackknife blocks; p-values of the two tests; level \\
$\beta(i)$, $\gamma(i)$ & medial correlation and Gaussian-copula correlation of the distances from point $i$ \\
$\mu_i(j)$, $\nu^F_j$ & smallest $k$ at which point $i$ keeps $j$; chance-corrected overlap of the $j$ farthest points \\
$\theta$, $\lambda$ & a method parameter along a path; rank weight of SR-tSNE2 \\
\bottomrule
\end{tabular}
\end{table}

\begin{table}[!htbp]
\centering\scriptsize
\caption{SR-tSNE2 has the highest or tied-highest $L$ on all twelve datasets; PCA has the highest $G$ on six and PaCMAP on four. $L$ (left) and $G$ (right) of seven methods on the twelve large datasets, the four applications first; bold: best on the dataset. t-SNE$_{200}$: perplexity 200.}\label{tab:largeLG}
\setlength{\tabcolsep}{2.6pt}
\begin{tabular}{@{}l*{7}{c}@{\hspace{8pt}}*{7}{c}@{}}
\toprule
& \multicolumn{7}{c}{$L$} & \multicolumn{7}{c}{$G$} \\
\cmidrule(lr){2-8}\cmidrule(l){9-15}
Dataset & PCA & NPCA & t-SNE & t-SNE$_{200}$ & SR-tSNE2 & UMAP & PaCMAP & PCA & NPCA & t-SNE & t-SNE$_{200}$ & SR-tSNE2 & UMAP & PaCMAP \\
\midrule
\csname @@input\endcsname tab_large_LG_rows
\bottomrule
\end{tabular}
\end{table}

\begin{table}[!htbp]
\centering\small
\caption{The twelve large datasets span 50,000 to 1.38 million points and 50 to 2,000 dimensions. $d$: dimension of the reference representation.}\label{tab:largedata}
\begin{tabular}{@{}llrrl@{}}
\toprule
Dataset & Domain & $n$ & $d$ & Representation \\
\midrule
\csname @@input\endcsname tab_large_data_rows
\bottomrule
\end{tabular}
\end{table}

\begin{table}[!htbp]
\centering\footnotesize
\caption{On the Swiss roll and the S-curve the neighbor embeddings are far more global than PCA. $L/G$ on the three synthetic manifolds, whose reference is the geodesic distance; mean over ten seeds for methods marked $^\dagger$ and for SR-tSNE2. Bold: best $L$ and best $G$ on each manifold.}\label{tab:synth}
\begin{tabular}{@{}lccc@{}}
\toprule
& Swiss roll & S-curve & Tree \\
\midrule
PCA & .41/.24 & .45/.29 & .10/.53 \\
NPCA & .41/.24 & .32/.13 & .07/.47 \\
KPCA & .41/.26 & .52/.52 & .01/.01 \\
LLE & .63/.60 & .55/.55 & .33/.57 \\
Isomap & .88/\textbf{.98} & .80/\textbf{.90} & .22/\textbf{.60} \\
t-SNE, PCA init & .85/.84 & .78/.85 & \textbf{.51}/.55 \\
SR-tSNE2 & \textbf{.93}/.81 & \textbf{.82}/.84 & .51/.56 \\
UMAP$^\dagger$ & .77/.85 & .72/.85 & .42/.42 \\
PHATE$^\dagger$ & .56/.73 & .56/.75 & .32/.25 \\
PaCMAP$^\dagger$ & .79/.66 & .74/.81 & .40/.52 \\
TriMAP$^\dagger$ & .72/.57 & .69/.71 & .40/.57 \\
\bottomrule
\end{tabular}
\end{table}

\begin{table}[!htbp]
\centering\footnotesize
\setlength{\tabcolsep}{4pt}
\caption{SR-tSNE2 never lowers $L$, raises it by up to 0.056, and moves $G$ by at most 0.0005. Our methods against the methods they modify on the twelve large datasets, $L/G$. $\Delta$: SR-tSNE2 minus the t-SNE map it refines.}\label{tab:own}
\begin{tabular}{@{}lcccccc@{}}
\toprule
Dataset & PCA & NPCA & t-SNE & SR-tSNE2 & $\Delta L$ & $\Delta G$ \\
\midrule
\csname @@input\endcsname tab_own_rows 
\bottomrule
\end{tabular}
\end{table}

\begin{table}[!htbp]
\centering\footnotesize
\caption{Raising t-SNE's perplexity from 5 to 500 trades $L$ (0.363 to 0.180) for $G$ (0.189 to 0.269); no setting reaches PCA's $G$ of 0.318. Local fidelity, global fidelity, the standard deviation of $G$ over three runs (t-SNE paths: one PCA and two random initializations), and energy (J, CPU time at 15 W per core) along each parameter path on 60,000 MNIST digits; means over three runs. PCA has $L=0.008$ and $G=0.318$.}\label{tab:sweep}
\begin{tabular}{@{}llrrrr@{}}
\toprule
Path & Value & $L$ & $G$ & sd $G$ & J \\
\midrule
\csname @@input\endcsname tab_sweep_rows 
\bottomrule
\end{tabular}
\end{table}

\begin{table}[!htbp]
\centering\scriptsize
\setlength{\tabcolsep}{2.6pt}
\caption{t-SNE and SR-tSNE2 are declared local and PaCMAP global on the same seven large datasets, and NPCA global on three, with AG News at the level for t-SNE. Map A, the column head, against map B, PaCMAP or NPCA. Last pair, upper end only: $\Delta G=G^{\rm PCA}-G^{\rm NPCA}$ and its one-sided $p_G$, bold below 0.05. $p$: p-value of the declaration, bold when declared at level 0.05. $s^\times$: crossover, in percent of $n$. Each test runs on a uniform subsample of 5,000 points. On AG News $p=0.04999$ for t-SNE against NPCA, at the level; we count it as borderline.}\label{tab:pairs}
\begin{tabular}{@{}l*{12}{c}@{}}
\toprule
& \multicolumn{6}{c}{B: PaCMAP} & \multicolumn{4}{c}{B: NPCA} & \multicolumn{2}{c}{B: PCA} \\
\cmidrule(lr){2-7}\cmidrule(lr){8-11}\cmidrule(l){12-13}
& \multicolumn{2}{c}{UMAP} & \multicolumn{2}{c}{t-SNE} & \multicolumn{2}{c}{SR-tSNE2} & \multicolumn{2}{c}{t-SNE} & \multicolumn{2}{c}{SR-tSNE2} & \multicolumn{2}{c}{NPCA} \\
\cmidrule(lr){2-3}\cmidrule(lr){4-5}\cmidrule(lr){6-7}\cmidrule(lr){8-9}\cmidrule(lr){10-11}\cmidrule(l){12-13}
Dataset & $p$ & $s^\times$ & $p$ & $s^\times$ & $p$ & $s^\times$ & $p$ & $s^\times$ & $p$ & $s^\times$ & $\Delta G$ & $p_G$ \\
\midrule
\csname @@input\endcsname tab_pairs_rows
\bottomrule
\end{tabular}
\end{table}

\begin{table}[!htbp]
\centering\scriptsize
\setlength{\tabcolsep}{2.6pt}
\caption{On 50 independent subsamples of 5,000 points the test of t-SNE against PCA never rejects a true null, and its declaration does not change. $\Delta G$: full-data $G^{\rm PCA}-G^{\rm t\text{-}SNE}$. SD: standard deviation of the estimate over subsamples. SE: mean standard error by the block jackknife and by a test that treats points as independent. Size: rate at which each rejects a true null at level 0.05, local and global. Declared: share of subsamples with a declared reversal. $s^\times$: mean crossover (standard deviation).}\label{tab:sim}
\begin{tabular}{@{}lcccccccccc@{}}
\toprule
 & & & \multicolumn{2}{c}{SE of $\Delta G$} & \multicolumn{2}{c}{Size, jackknife} & \multicolumn{2}{c}{Size, independent} & & \\
Dataset & $\Delta G$ & SD & jackknife & independent & $L$ & $G$ & $L$ & $G$ & Declared & $s^\times$ \\
\midrule
\csname @@input\endcsname tab_sim_rows 
\bottomrule
\end{tabular}
\end{table}

\begin{table}[!htbp]
\centering\small
\caption{Which map to use, by the LG reversal test against PCA on the four applications; the local map is t-SNE unless the row names another. The crossover size is $s^\times n$, the number of points at the crossover, with its 95\% jackknife interval.}\label{tab:lguse}
\begin{tabular}{@{}>{\raggedright\arraybackslash}p{2.5cm}>{\raggedright\arraybackslash}p{3.4cm}>{\raggedright\arraybackslash}p{3.4cm}>{\raggedright\arraybackslash}p{3.6cm}@{}}
\toprule
Dataset & Question about fewer points: use the local map & Question about more points: use the global map & Crossover size \\
\midrule
Pathology-100K & the tiles that resemble a tumor tile & the layout of the nine tissues & 10,040 tiles (4,210--15,870) \\
Pathology-7K & the tiles that resemble a tumor tile & the layout of the nine tissues & 776 tiles (387--1,165) \\
MS MARCO & the passages near a query and its retrieved set (t-SNE) & the topics of the memory and their distances (PCA) & 150,400 passages (79,100--221,700), t-SNE against PCA \\
8-K & the filings near a new event (UMAP) & the topics of the memory (PCA) & 427 filings (226--628), UMAP against PCA \\
SEA-AD & \multicolumn{3}{p{10.4cm}}{t-SNE for the neighbors of a cell and for subtypes; the global gap of 0.044 between PCA and t-SNE stays within its sampling error ($p=0.16$), and PaCMAP has the highest $G$, 0.417, for the distances between cell types} \\
Instacart & \multicolumn{3}{p{10.4cm}}{against PCA, t-SNE is more local by 0.080 and more global by 0.021; PaCMAP has the highest $G$, 0.268, for the layout of segments} \\
Online Retail & \multicolumn{3}{p{10.4cm}}{t-SNE leads at both ends: it is more local by 0.239 and more global by 0.088} \\
\bottomrule
\end{tabular}
\end{table}

\FloatBarrier
\sitablesend
\section{Sensitivity of $G$ to its band}\label{app:sensitivity}

$G$ changes little when its band $0.3\le k/n\le0.5$ is replaced. Supplementary Table~\ref{tab:sens} replaces the band by alternatives on the 84 maps of the twelve large datasets. It reports three quantities: the rank correlation with $G$, the agreement of the method rankings within each dataset, and the rank correlation with the two established global scores. All variants are read from the same stored profiles, so the comparison involves no new maps.

\begin{table}[ht]
\centering\small
\setlength{\tabcolsep}{4.5pt}
\caption{$G$ changes little with its band (84 maps of the twelve large datasets). Default: correlation with $G$ across all maps. Kendall $\tau$: agreement of the method rankings within a dataset, mean and minimum. Distance $\rho$ and triplets: correlation with the distance rank correlation and with random-triplet accuracy.}\label{tab:sens}
\begin{tabular}{@{}lccccc@{}}
\toprule
Variant & Default & $\tau$ mean & $\tau$ min & Distance $\rho$ & Triplets \\
\midrule
\csname @@input\endcsname tab_sens_rows 
\bottomrule
\end{tabular}
\end{table}

\runhead{Global fidelity.} The bands $0.2$--$0.5$ and $0.4$--$0.5$ correlate at 0.98 with $G$ and the midpoint at 0.96, and these scores correlate at 0.80 to 0.94 with the distance rank correlation and 0.91 to 0.98 with random-triplet accuracy. Narrower bands at the top of the profile agree more closely with the established global scores. Within a dataset they rank the methods with mean $\tau$ 0.89 for $0.4$--$0.5$ and 0.79 for the midpoint alone. These bands average fewer values of $k$, and the methods on a large dataset often lie within 0.01 of each other in $G$. The middle band $0.1\le k/n\le0.2$ ranks the methods differently from $G$ (mean $\tau$ 0.15) and agrees with the references at 0.57 and 0.70. The default band is the widest that stays at the upper end.

\section{Calibration of the local test}\label{app:localcal}

On independent subsamples the local test rejects a true null at no more than 0.020 (Supplementary Note~\ref{sec:sim}), and this is its calibration. Independent subsamples reproduce the drawing of points, which the centered null of Results, built for the global test, holds fixed. The local declarations rest on $p_L\le1.3\times10^{-5}$ (Figure~\ref{fig:large}b), far below any of these rates.

\section{The local test on all points}\label{app:localfull}

The local test also runs on all points. Per-point $L(i)$ uses the exact 10 nearest neighbors in the data and in the map, so its mean is the exact $L$ of Supplementary Table~\ref{tab:largeLG}. The 20 $k$-means blocks are fitted on 50,000 points and assigned to all, and the jackknife is the one described in Methods. Against the t-SNE map it refines, SR-tSNE2 gains in $L$ on nine large datasets, by 0.010 to 0.056 with $p\le3.7\times10^{-5}$. On CIFAR-10, Speech Commands and AG News the gain is 0.0002 to 0.0007, within its sampling error. Under the centered null the twelve tests reject at 0.053 on average at level 0.05, and from 0.014 to 0.148. On the 5,000-point subsample used by the test, the 10 nearest neighbors stand for about the 100 to 2,800 nearest in the full data, and there t-SNE leads SR-tSNE2 by 0.001 to 0.049. The rank refinement thus acts at the exact neighbor scale it optimizes.

\section{Number of blocks}\label{app:blocks}

The declarations of t-SNE against PCA change little with the number of blocks $C$. With $C=10$, 20 and 50 the declaration is the same on nine of the twelve large datasets, and the crossover does not depend on $C$. With $C=10$ six of the seven declarations of Figure~\ref{fig:large}a hold, and on MNIST $p$ rises to 0.061. With $C=50$ all seven hold, and CIFAR-10 and Renal-Cortex join them with $p=0.049$ and 0.042. More blocks give smaller standard errors, so $C=10$ gives the most cautious test and $C=20$ lies between. Under the centered null the global test rejects at 0.059, 0.046 and 0.043 on average for $C=10$, 20 and 50.

\section{Refitting the map}\label{app:refit}

A refit of the map moves the global difference far less than the sampling of points does. On MNIST we refitted t-SNE five more times from random starts and tested each map against PCA as in Figure~\ref{fig:large}. Over the six maps $G^{\rm PCA}-G^{\rm t\text{-}SNE}$ lies between 0.104 and 0.130, with standard deviation 0.010 across maps. Its jackknife standard error is 0.045 to 0.074, about five to seven times larger. The crossover lies between 0.180 and 0.197. The reversal is declared for four of the six maps, and for the other two $p=0.051$ and 0.061, close to the level, as the $p$ of 0.012 for the first map leads one to expect.

\end{document}